\documentclass{article}
\usepackage{iclr2027_conference,times}
\usepackage[T1]{fontenc}
\usepackage{microtype}
\usepackage{hyperref}
\hypersetup{hidelinks,hypertexnames=false}
\usepackage{xurl}
\usepackage{booktabs}
\usepackage{array}
\usepackage{graphicx}
\usepackage{amsmath}
\usepackage{float}
\usepackage{wrapfig}
\usepackage{needspace}
\usepackage{placeins}
\usepackage{etoolbox}

\AtBeginEnvironment{figure}{\setlength{\abovecaptionskip}{4pt}\setlength{\parskip}{0pt}}

\makeatletter
\def\input@path{{./}{sections/}{tables/}}
\makeatother
\graphicspath{{./}}

\iclrfinalcopy
\makeatletter
\patchcmd{\@maketitle}{Published as a conference paper at ICLR 2027}{Preprint}{}{\PackageError{arxiv}{Header patch failed}{}}
\makeatother

\author{Haitong Jiang, Chunlin Liu, Yile Wang$^{*}$, Yuhong Feng$^{*}$\\
College of Computer Science and Software Engineering, Shenzhen University\\
{\normalfont\small $^{*}$Corresponding authors: \texttt{\{wangyile,yuhongf\}@szu.edu.cn}.}}
\hypersetup{pdfauthor={Haitong Jiang, Chunlin Liu, Yile Wang, Yuhong Feng}}

\begin{document}
\raggedbottom
\title{Exact Feedback Is Not Control: Evaluating Text-based Closed-Loop Revision in LLMs}
\maketitle

\begin{abstract}
Closed-loop revision is increasingly adopted in large language model applications to iteratively reduce mismatches between generated outputs (e.g., a paragraph) and user targets (e.g., required paragraph length). However, revision failures may stem from incomplete feedback or models failing to act effectively on correct feedback, making it difficult to systematically assess the revision capabilities of LLMs. In this work, we propose a fixed-budget revision protocol in which deterministic verifiers report all remaining violations across three constraint families: exact-length, lexical, and compositional. This holds feedback correctness and completeness fixed, allowing us to isolate and evaluate model-side closed-loop revision. We conduct experiments on 19 mainstream open-source and closed-source models, including Llama 3.1 8B and GPT-5.6 Sol, and observe large cross-model and cross-constraint differences in final joint success, with controller-level mean final joint success ranging from 17.4\% to 99.8\%. Substantial cross-model gaps persist when the same initial draft is used. In controlled experiments, we find reproducible model-specific differences in how exact feedback is translated into revisions, while post-training and model scale reshape these responses without consistently bringing them closer to exact correction. Interestingly, across all constraint families, failure trajectories often repeat earlier outputs, and prior recurrence is associated with substantially lower subsequent recoverability. Finally, matched-state interventions across constraint families show that removing earlier dialogue, with the current draft and feedback fixed, changes recurrence escape without reliably improving final success; effects depend on the model, task, and trigger-state composition. Overall, exact feedback makes revision errors observable, but does not make the closed loop reliable. Code and reproduction instructions are available\footnote{\url{https://github.com/kevinjiang0121-cyber/exact-feedback-code}}.
\end{abstract}

\section{Introduction}

Feedback-driven revision is increasingly used in large language model (LLM) systems to bring
generated outputs closer to user requirements. Such systems revise behavior
across trials \citep{shinn2023reflexion}, modify code using execution results
\citep{yang2024sweagent}, replan from environmental observations
\citep{huang2023innermonologue}, and refine candidate programs using evaluator
scores \citep{novikov2025alphaevolve}. In each setting, feedback identifies a
gap between the current output and the target, and LLMs must
decide how to act on it repeatedly and ultimately complete the task successfully.

However, revision can be unreliable due to a lack of reliable external feedback in
reasoning \citep{huang2024large} and inaccurate model-generated feedback by itself \citep{olausson2024selfrepair}.
Studies show that models can persist with
incorrect answers despite critiques \citep{jiang2025feedback} and
surveys further emphasize the role of feedback quality \citep{kamoi2024when}. When feedback itself may be unreliable, poor revision outcomes do not reveal whether the bottleneck lies in the feedback or in the model’s response to it. This leaves open how reliably LLMs can act on feedback over repeated revisions when that feedback is complete and correct.

\begin{figure}[t!]
\centering
\includegraphics[width=\textwidth,trim=0 22.63bp 0 0,clip]{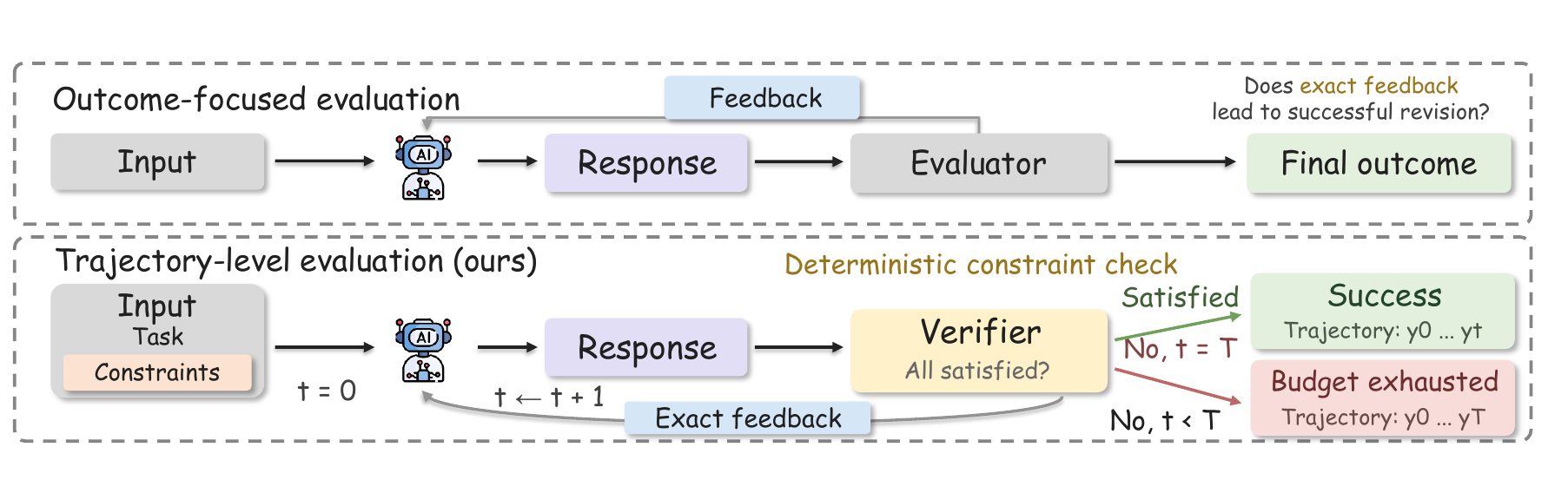}
\caption{Two views of LLM revision: outcome-focused evaluation (top) and trajectory-level evaluation under exact feedback (bottom, this work).}
\label{fig}
\end{figure}

In this work, we propose a trajectory-level framework to evaluate fixed-budget closed-loop revision in LLMs, as illustrated in Figure~\ref{fig}. We study revision under explicit, deterministically checkable constraints and use verifiers that report all remaining violations after each revision. The same frozen rules generate feedback and determine final success, making feedback correctness and completeness controlled experimental conditions and allowing us to isolate model-side revision.

In particular, we build a reusable evaluation suite spanning three text-based constraint families—exact-length, lexical, and compositional—and use it to evaluate 19 mainstream open- and closed-source models. We first examine cross-model and cross-task variation, including fixed-draft comparisons that hold the starting response constant. We then characterize model-specific feedback-to-action responses under controlled correction requests, and test whether these patterns replicate on unseen texts and remain informative during independent multi-round revision. Finally, we analyze recurrent failure trajectories and use matched-state history interventions to examine how retained dialogue shapes subsequent editing behavior and final success.

Overall, our contributions and findings can be summarized as follows:

\begin{itemize}
\item \textbf{An auditable framework and evaluation suite for isolating model-side revision.}
We introduce a fixed-budget request--model--verifier loop for closed-loop revision. Deterministic verifiers report all remaining violations and use the same frozen rules to determine final success, making feedback correctness and completeness controlled experimental conditions. The accompanying evaluation suite provides frozen cases, trajectory-level records, and generation and analysis pipelines for reproducible evaluation.

\item \textbf{A systematic characterization of closed-loop revision behavior.}
We evaluate 19 model configurations across three text-based constraint families and find substantial cross-model and cross-constraint variation. Large revision gaps persist from the same starting responses, showing that they are not simply inherited from differences in initial generation.

\item \textbf{A controlled analysis of feedback-to-action responses.}
By holding drafts fixed and varying exact correction requests, we identify reproducible, model-specific response patterns and show that post-training and model scale reshape these behaviors in different ways. These response patterns remain informative during independent multi-round revision.

\item \textbf{A trajectory-level analysis of output recurrence and history effects.}
Failed revision trajectories frequently revisit earlier outputs across all three constraint families. Matched history interventions substantially alter escape from recurrence and subsequent editing behavior, while producing much smaller and heterogeneous changes in final success.

\end{itemize}

\section{Related Work}
\label{sec:related-work}

\paragraph{Iterative refinement and self-correction.}
Self-Refine alternates critique and revision using the same model, without
additional training \citep{madaan2023selfrefine}. CRITIC grounds critiques
in external tool feedback \citep{gou2023critic}, while SCoRe trains
multi-turn self-correction through reinforcement learning
\citep{kumar2024training}. These approaches improve revision through
different feedback sources or changes to the revision policy.
Their effectiveness depends on the setting: reasoning studies find that
self-correction without external feedback can fail to improve answers
\citep{huang2024large}, and code-repair studies identify feedback quality
as a limitation \citep{olausson2024selfrepair}. A broader survey likewise
emphasizes the conditions under which feedback supports correction
\citep{kamoi2024when}.
Closer to our question, Feedback Friction finds that models can persist with incorrect answers despite targeted critiques generated with access to ground-truth answers \citep{jiang2025feedback}. RefineBench reports substantial cross-model variation in refinement performance \citep{lee2026refinebench}.
These studies leave open whether cross-model refinement gaps persist when the starting draft and complete, exact feedback are held fixed, and whether such gaps reflect reproducible differences in how models translate the same feedback into revisions.

\paragraph{Exact verification and constrained revision.}
Constrained decoding enforces lexical and logical requirements during
generation \citep{lu2021neurologic}, while length-control methods target
outputs of a prescribed length \citep{li2024ruler}.
We study how a revision policy acts on exact feedback under standard decoding.
Reliable verification has also been used for repeated correction.
On arithmetic, graph coloring, and planning, \citet{stechly2025verification}
compare sound external feedback at different levels of detail with repeated
sampling. DeCRIM decomposes multi-constraint instructions and uses critiques
to guide revision, including rule-based feedback on IFEval
\citep{ferraz2024decrim}.
Our controlled experiments hold drafts fixed and vary accurate correction
requests, measuring the resulting edits from the input--output perspective
of system identification \citep{ljung1999system}. We test whether these
responses replicate on held-out texts and remain informative during
independent revision, then examine failure trajectories and history effects
under complete violation reports.

\paragraph{Closed-loop behavior and history.}
Trajectory studies track answers repeatedly in question answering
\citep{jiang2025feedback}, correctness transitions during self-correction
\citep{liu2026feedbackcontrol}, and textual fixed points during repeated
abstract polishing \citep{wu2026relaxation}.
Our analysis follows repeated full outputs while explicit constraints
remain unsatisfied, connecting trajectory behavior to constraint satisfaction
as in closed-loop verification \citep{wang2020neural}.
History also matters when instructions are gradually revealed: models
can rely too heavily on earlier answers \citep{laban2025llms}.
We test history's role after task requirements are fully specified.
By holding the current draft and verifier report fixed, we distinguish
the effect of earlier dialogue from differences in the text being revised
or the feedback it receives. We measure intervention effects on the next
revision and final success separately, testing whether gains in immediate
behavior are accompanied by gains in task completion.

\section{A Unified Closed-Loop Revision Protocol}
\label{sec:assay}

\paragraph{Problem definition.}
Given a task $x$ and an initial draft $y_0$, we study whether an LLM
can satisfy a set of verifiable constraints $\mathcal{C}_x$ within
a budget of $T$ revisions. Joint success requires every constraint to hold
in the same draft:
\begin{equation}
J(x,y)=\prod_{c\in\mathcal{C}_x}c(y),
\label{eq:revision-objective}
\end{equation}
where each constraint is a binary check: $c(y)=1$ if draft $y$
satisfies $c$, and $c(y)=0$ otherwise. Thus $J(x,y)=1$ indicates
joint success.

Let $y_t$ denote the draft after $t$ revisions. The verifier $V_k$ for constraint family $k$ checks
$y_t$ and reports all remaining violations. If the draft is unsuccessful
and the budget has not been exhausted, model $m$ produces $y_{t+1}$
according to its revision policy $\pi_m$:
\begin{equation}
V_k(x,y_t)=(J_t,v_t),
\qquad
 y_{t+1}\sim\pi_m(\cdot\mid x,y_t,v_t,h_t),
\label{eq:protocol}
\end{equation}
where $J_t=J(x,y_t)$ is the success indicator, $v_t$ is the violation
report, and $h_t$ contains retained earlier drafts and feedback.
The policy $\pi_m$ specifies the distribution over revised texts given
the task, current draft, feedback, and history, under the model's fixed
tokenizer, chat template, and decoding configuration. This
configured model is called a ``controller'' and also a ``reviser'' when we describe its role
in revising a supplied draft. Each model generates its own initial draft $y_0$ in the main evaluation; fixed-draft comparisons also use drafts generated by other models.

\begin{figure}[t!]
  \centering
  \setlength{\abovecaptionskip}{2pt}
  \includegraphics[width=\textwidth]{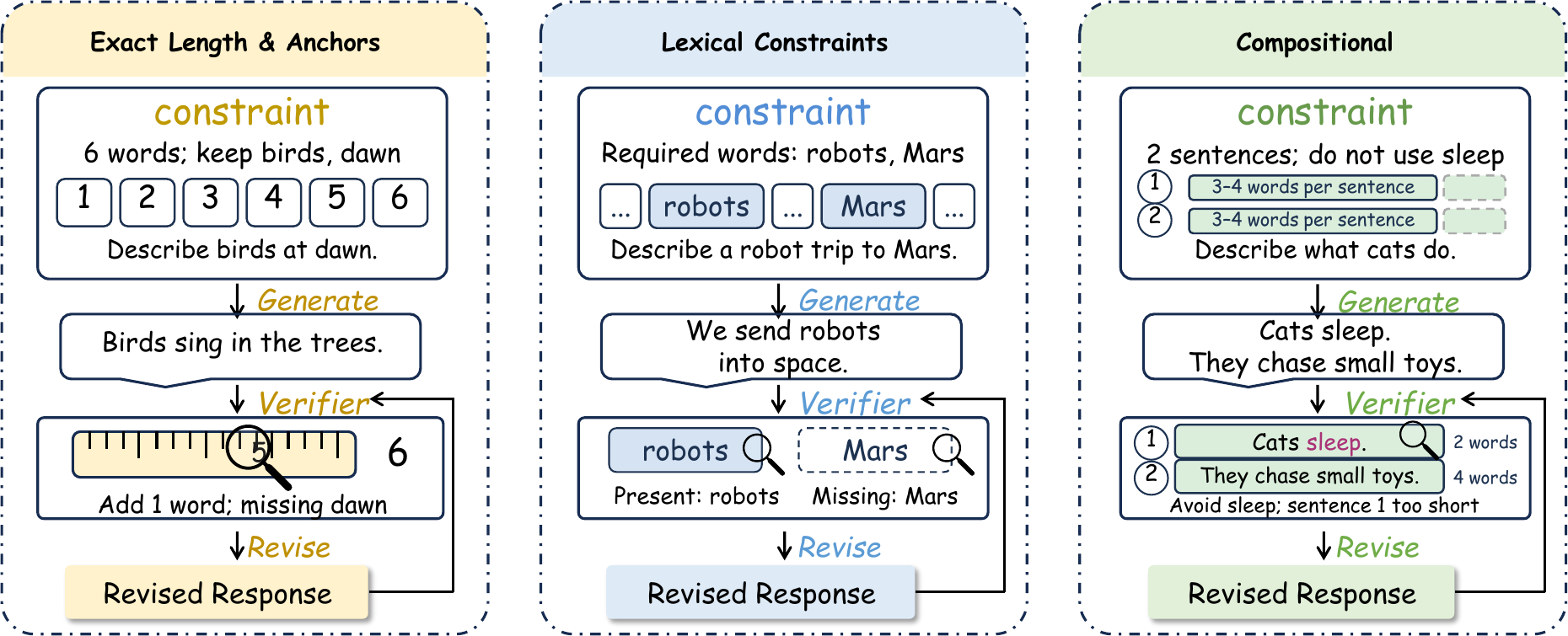}
  \caption{Illustration of exact-length, lexical, compositional constraints and the shared revision loop.}
  \label{fig:assay-overview}
\end{figure}

\paragraph{Constraint families.}
We apply the protocol to three families of verifiable constraints, as Figure~\ref{fig:assay-overview} shows.
The \emph{exact-length constraints} require both the requested word count and a
fixed set of content anchors: specified words or phrases that must be retained.
The feedback reports the current and target word counts, the required addition
or deletion, and any missing anchors. The \emph{lexical constraints} require
specified words to appear, occupy specified positions, or end specified
sentences, with word- or sentence-count requirements where applicable. The \emph{compositional constraints} combine sentence count with
forbidden words or per-sentence length bounds. The latter two constraint families together use five
structures from COLLIE \citep{yao2023collie}; each case includes a human-written
response that satisfies all its constraints. We fix the rules for word counting, sentence splitting, and matching,
as well as the feedback fields, so that every violation report is reproducible.
Detailed constraint definitions are given in Appendix~\ref{app:verifiers}.

\paragraph{Loop outcomes.}
We measure whether the loop succeeds and how many revisions it takes.
The capture time is the index of the first draft that satisfies all constraints:
\begin{equation}
  T_{\mathrm{cap}}=\min\{0\leq t\leq T:J_t=1\},
  \label{eq:capture-time}
\end{equation}
where we allow up to $T$ revisions in total, stopping at the first successful draft or when the budget is exhausted ($T_{\mathrm{cap}}=\infty$).
Final joint success (shortened to final success) is
$\mathbf{1}[T_{\mathrm{cap}}\leq T]$, where $\mathbf{1}[\cdot]$ equals
one when its condition holds and zero otherwise.

We fixed $T=8$ in the main experiments to allow repeated correction while limiting inference cost and dialogue growth. Each model receives the same maximum number of revisions while token costs can differ. Eight rounds do not imply convergence and Appendix~\ref{app:budget-sensitivity} reports an audit that continues selected failures to round 32. To describe how revision progresses, we also track error reduction and output repetition. \textit{Contraction} means a decrease in the violation measure for the task's constraint family. An unsuccessful output \textit{recurs} when it is byte-identical to an earlier draft.

\paragraph{Evaluation design.}
We evaluate 480 cases per constraint family. Exact-length targets and
anchors are derived from human-written references and prompts. Lexical
and compositional cases come from COLLIE~\citep{yao2023collie}. Cases are shared across models
within each family, but differ across families. We test 19 model
configurations---12 local checkpoints from seven model families and
seven API systems---under the eight-revision protocol, yielding 27,360
attempted trajectories. Success rates use protocol-complete records;
paired comparisons use cases available for all models being compared.
We use source-stratified bootstrap intervals with 20,000 resamples,
exact McNemar tests \citep{mcnemar1947sampling} with Holm correction
\citep{holm1979sequentially}, and case-clustered interaction
models for the main evaluations. Data and statistical details are given
in Appendix~\ref{app:reproducibility}.

\begin{figure}[!t]
\centering

\setlength{\abovecaptionskip}{2pt}
\includegraphics[width=\textwidth]{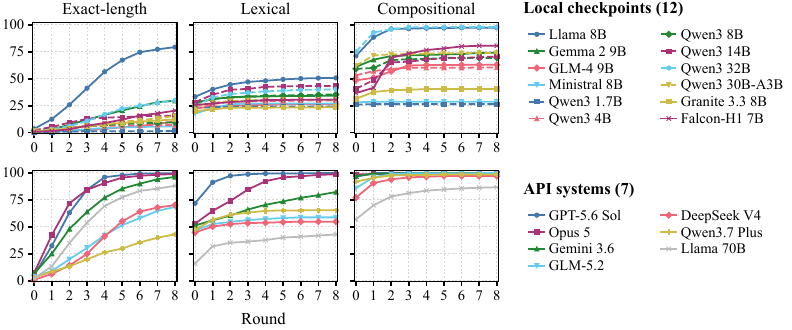}
\vspace{-6pt}
\caption{Cumulative joint success across three constraint families. Round 0 is the initial draft. Full model names appear in Table~\ref{tab:all-model-assay-matrix}.}
\label{fig:revision-round-profiles}
\end{figure}

\section{Revision Performance and Feedback-to-Action Responses}
\label{sec:closed-loop-gap}

\subsection{Revision Performance across Rounds and Models}
Table~\ref{tab:all-model-assay-matrix} shows substantial variation in final joint success even under exact feedback. API systems perform better on average than local checkpoints, but large differences remain within both groups, with controller-level mean success ranging from 17.4\% to 99.8\%. Performance also varies strongly across constraint families: exact-length and lexical tasks remain challenging for many models, whereas compositional tasks generally achieve higher success.

\newpage
\begingroup
\setlength{\columnsep}{8pt}
\hyphenpenalty=5000
\exhyphenpenalty=5000
\emergencystretch=1em
\setlength{\intextsep}{5pt}
\begin{wraptable}{r}{0.46\textwidth}
\begin{minipage}[t]{\linewidth}
\vspace{0pt}
  \centering
  \scriptsize
  \setlength{\tabcolsep}{1.5pt}
  \renewcommand{\arraystretch}{0.84}

  \begin{tabular*}{\linewidth}{@{\extracolsep{\fill}}lrrrr@{}}
    \toprule
    \textbf{Model} & \textbf{Exact} & \textbf{Lex.} & \textbf{Comp.} & \textbf{Mean $\pm$ SD} \\
    \midrule

\multicolumn{5}{@{}l}{\emph{Local checkpoints}} \\
    \textsuperscript{1} Llama 3.1 8B       & 79.4 & 50.8 & 97.3 & 75.8 $\pm$ 23.5 \\
    \textsuperscript{2} Qwen3 32B          & 30.0 & 40.2 & 97.9 & 56.0 $\pm$ 36.6 \\
    \textsuperscript{3} Gemma 2 9B         & 29.6 & 34.2 & 74.2 & 46.0 $\pm$ 24.5 \\
    \textsuperscript{4} Falcon-H1 7B       & 20.4 & 30.6 & 80.6 & 43.9 $\pm$ 32.2 \\
    \textsuperscript{5} Qwen3 14B          & 15.6 & 43.3 & 70.6 & 43.2 $\pm$ 27.5 \\
    \textsuperscript{6} Qwen3 30B-A3B      & 15.4 & 30.6 & 74.0 & 40.0 $\pm$ 30.4 \\
    \textsuperscript{7} Qwen3 8B           &  9.6 & 35.4 & 69.4 & 38.1 $\pm$ 30.0 \\
    \textsuperscript{8} GLM-4 9B           & 10.4 & 26.3 & 62.9 & 33.2 $\pm$ 26.9 \\
    \textsuperscript{9} Qwen3 4B           &  7.1 & 29.8 & 60.4 & 32.4 $\pm$ 26.7 \\
    \textsuperscript{10} Granite 3.3 8B     & 12.3 & 23.8 & 40.4 & 25.5 $\pm$ 14.1 \\
    \textsuperscript{11} Ministral 8B       &  5.8 & 27.1 & 28.8 & 20.6 $\pm$ 12.8 \\
    \textsuperscript{12} Qwen3 1.7B         &  1.5 & 24.2 & 26.5 & 17.4 $\pm$ 13.8 \\
    \emph{Local mean ($n=12$)} & \emph{19.8} & \emph{33.0} & \emph{65.2} & \emph{39.3 $\pm$ 23.4} \\

    \midrule
    \multicolumn{5}{@{}l}{\emph{API systems}} \\
    \textsuperscript{13} GPT-5.6 Sol         & 99.6 & 99.8 & 100.0 & 99.8 $\pm$ 0.2 \\
    \textsuperscript{14} Claude Opus 5       & 99.0 & 98.8 &  99.2 & 99.0 $\pm$ 0.2 \\
    \textsuperscript{15} Gemini 3.6 Flash    & 96.2 & 82.3 &  99.8 & 92.8 $\pm$ 9.2 \\
    \textsuperscript{16} GLM-5.2             & 68.5 & 58.8 &  99.8 & 75.7 $\pm$ 21.4 \\
    \textsuperscript{17} DeepSeek V4 Flash   & 70.1 & 54.8 &  97.1 & 74.0 $\pm$ 21.4 \\
    \textsuperscript{18} Llama 3.1 70B       & 88.1 & 42.9 & 86.7 & 72.6 $\pm$ 25.7 \\
    \textsuperscript{19} Qwen3.7 Plus        & 43.0 & 65.4 &  98.5 & 69.0 $\pm$ 27.9 \\
    \emph{API mean ($n=7$)} & \emph{80.6} & \emph{71.8} & \emph{97.3} & \emph{83.3 $\pm$ 12.9} \\
    \midrule
    \textbf{Overall mean ($n=19$)} & \textbf{42.2} & \textbf{47.3} & \textbf{77.1} & \textbf{55.5 $\pm$ 18.8} \\
    \bottomrule
  \end{tabular*}
  \setlength{\abovecaptionskip}{4pt}
  \caption{Final joint success (\%); the last column reports the mean and sample SD across the three constraint families.}
  \label{tab:all-model-assay-matrix}
\end{minipage}

\par\vspace{8pt}
\begin{minipage}[t]{\linewidth}
\vspace{0pt}
\makeatletter\def\@captype{table}\makeatother
\centering\scriptsize
\renewcommand{\arraystretch}{0.84}
\setlength{\tabcolsep}{1.5pt}
\begin{tabular*}{\linewidth}{@{\extracolsep{\fill}}llr@{}}
\toprule
\textbf{Draft Source} & \textbf{Reviser} & \textbf{Success} \\
\midrule
Llama 3.1 8B & Llama 3.1 8B & 76.2 \\
 & Qwen3 14B & 17.1 \\
 & Gemma 2 9B & 33.8 \\
 & GLM-4 9B & 12.9 \\
\midrule
Qwen3 14B & Llama 3.1 8B & 84.6 \\
 & Qwen3 14B & 15.8 \\
\midrule
Gemma 2 9B & Llama 3.1 8B & 81.2 \\
 & Gemma 2 9B & 25.4 \\
\midrule
GLM-4 9B & Llama 3.1 8B & 85.4 \\
 & GLM-4 9B & 8.3 \\
\bottomrule
\end{tabular*}
\setlength{\abovecaptionskip}{4pt}
\caption{Final joint success (\%) from fixed initial drafts on exact-length tasks.}
\label{tab:fixed-draft}
\end{minipage}

\end{wraptable}
\noindent At the same time, this overall local--API separation is not uniform. Llama 3.1 8B, for example, reaches a mean success rate of 75.8\%, matching or outperforming four of the seven API systems despite being a local 8B checkpoint. Overall, \textbf{exact feedback does not eliminate substantial model- and task-dependent variation, and closed-loop revision performance does not follow a simple ordering by deployment category or nominal model scale}.

Figure~\ref{fig:revision-round-profiles} shows cumulative joint success over revision rounds $0$--$8$ and reveals distinct recovery profiles across both tasks and models. \textbf{Exact-length tasks benefit most visibly from repeated revision}: several controllers continue to improve well into the later rounds, although the rate of recovery differs substantially across models. Some systems approach saturation after only a few revisions, while others, including Llama 3.1 8B and several API models, continue making meaningful gains near the end of the budget. Lexical tasks show a different pattern: most improvements occur in the early rounds and then diminish, leaving many models at a relatively stable plateau. Compositional tasks are more front-loaded, with many controllers already performing strongly after the first few revisions and showing little additional gain thereafter. These curves also reveal that the same revision budget has very different value across controllers: some models continue recovering from earlier failures, whereas others stop improving much earlier. Overall, \textbf{models differ not only in final success, but also in the speed, persistence, and timing of recovery over repeated revisions}. Detailed round-by-round results are reported in Appendix~\ref{app:budget-sensitivity}.

Are later revision gaps simply inherited from differences in initial generation ability? We test this on 240 exact-length cases, where local-model revision performance is most clearly separated. In these fixed-draft comparisons, different revisers receive the same task, initial draft, target, anchors, and revision budget, while retaining their native interfaces as in the main evaluation. Table~\ref{tab:fixed-draft} shows that, even from identical starting drafts, Llama 3.1 8B retains a 42.5--77.1 percentage-point advantage in final success across the six comparisons, all significant after Holm correction. Notably, Llama succeeds more often when revising each paired model's drafts than when revising its own. The advantage remains 45.3--77.1 points after excluding initially successful drafts (Appendix~\ref{app:crossover}). Overall, \textbf{large cross-model revision gaps persist even when the initial draft is held fixed, showing that they are not simply inherited from differences in initial generation}.

\par\WFclear
\endgroup

\subsection{Revision Responses Across Post-training and Model Scale}
\label{sec:response-curves}

Exact-length feedback lets us directly compare the requested word-count change with the actual edit. Because natural revision trajectories entangle correction requests with model-dependent drafts and histories, we hold each source draft fixed and vary only its target word count and corresponding verifier feedback. These controlled one-step revisions characterize feedback-to-action responses, while the original multi-round trajectories are reserved for testing whether the resulting response curves generalize to naturally occurring revision states.

For target $N$ and word count $W(y_t)$, we define the signed error $d_t$ and actual word-count change $a_t$:
\begin{equation}
d_t=W(y_t)-N,\qquad
a_t=W(y_{t+1})-W(y_t),\qquad
d_{t+1}=d_t+a_t.
\label{eq:dynamics}
\end{equation}
The requested change is $c=-d_t$: positive values request additions and
negatives request deletions. An exact correction has $a_t=c$, correct direction means $a_td_t<0$, and error expansion means
$|d_{t+1}|>|d_t|$. Zero action means no change in word count
($a_t=0$), though the text may change.

\begin{figure}[!t]
\centering
\includegraphics[width=\textwidth]{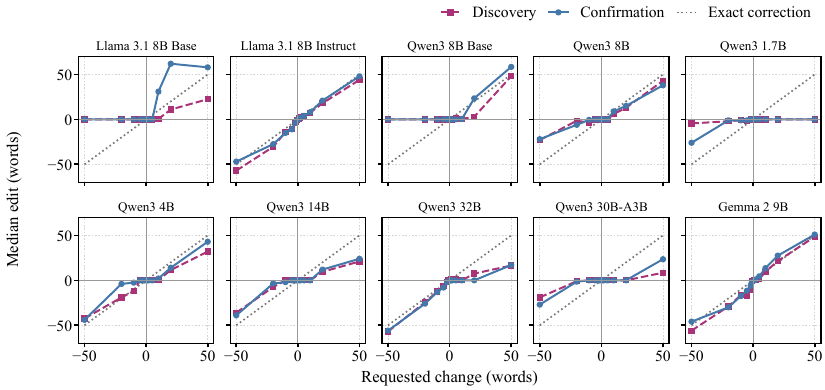}
\caption{Revision curves from discovery and confirmation texts for all ten checkpoints, spanning post-training conditions and model sizes. Dotted lines mark exact correction, where the median word-count change equals the requested change.}
\label{fig:policy-identification}
\end{figure}

We estimate each model's response curve \(g_m(c)\) as the median word-count change at each requested correction. Curves are estimated on 24 \emph{discovery} texts and evaluated without refitting on 48 disjoint \emph{confirmation} texts. Figure~\ref{fig:policy-identification} shows all ten checkpoints on the same texts and request grid. Discovery and confirmation curves correlate strongly for every checkpoint (Spearman \(\rho > 0.85\)), indicating that the response patterns replicate across unseen texts.

Figure~\ref{fig:policy-identification} also shows that post-training and model scale reshape revision responses in different ways. Both Llama 3.1 8B Base and Qwen3 8B Base exhibit near-zero responses over a range of deletion requests, but instruction tuning reshapes these behaviors differently. For Llama, instruction tuning largely removes this deletion-side dead zone and brings both directions closer to exact correction; for Qwen, it reduces error-expanding edits but also increases zero-action responses. Across the dense Qwen3 series from 1.7B to 32B, increasing model size does not bring responses uniformly closer to exact correction; for example, Qwen3 32B can over-correct deletion requests while still under-correcting additions. Gemma 2 9B provides a contrasting response profile, with median edits tracking the exact-correction line relatively closely in both directions. Overall, the curves reveal distinct response failures, including directional asymmetry, under-correction, and near-zero action. The Qwen3 30B-A3B MoE checkpoint provides a contrasting case, with zero median word-count change over tested requests from $-10$ to $+20$ words in both discovery and confirmation. Quantitative comparisons are in Appendix~\ref{app:controlled-interventions}.

Taken together, these results reveal systematic differences in how models translate exact feedback into revision actions. \textbf{Post-training can substantially reshape a shared Base-model failure pattern, but does so in checkpoint-specific ways, while increasing model scale does not consistently move responses toward exact correction}; models instead exhibit distinct patterns of under-correction, asymmetric editing, and near-zero action. Importantly, these differences extend beyond the controlled setting: across nine checkpoints, model-specific response curves reduce edit-prediction MAE on independent multi-round trajectories by 1.91 words relative to a shared curve, an 11.0\% relative reduction (95\% CI 8.5--13.9). Thus, \textbf{feedback-to-action responses are reproducible and remain informative as revision states evolve, but better local responses alone do not determine successful closed-loop revision}. Detailed comparisons and baselines are in Appendix~\ref{app:controlled-interventions}.

\section{Failure Dynamics and Targeted Interventions}
\label{sec:interventions}

\subsection{Output Recurrence across Constraints}
\label{sec:recurrence-law}

\begin{figure}[b!]
\centering
\includegraphics[width=\textwidth]{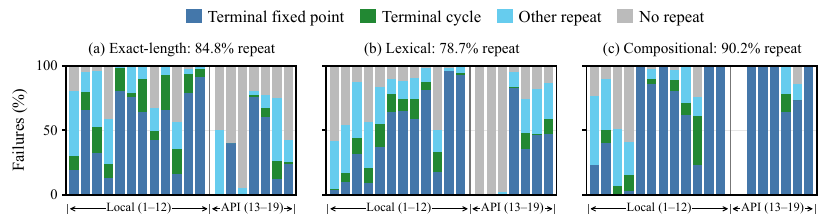}
\caption{Output recurrence conditional on failure. Each bar is normalized by that model's failure count within the constraint family. Bars follow the model order in Table~\ref{tab:all-model-assay-matrix}, grouped as local (1--12) and API (13--19); the empty GPT-5.6 Sol bar in (c) indicates no failures.}
\label{fig:failure-anatomy}
\end{figure}

\textbf{Failed trajectories frequently repeat complete outputs.}
Among failed trajectories, recurrence occurs in 84.8\% of 5,272 exact-length, 78.7\% of
4,805 lexical, and 90.2\% of 2,093 compositional trajectories, respectively.
In Figure~\ref{fig:failure-anatomy}, we compare complete outputs by exact byte equality and distinguish four
patterns: a \textbf{terminal fixed point}
ends with three identical drafts; a \textbf{terminal cycle} ends with two
identical consecutive blocks of $p\in\{2,3,4\}$ drafts, each containing
at least two distinct outputs; \textbf{other repeat} contains repetition
without either terminal pattern; and \textbf{no repeat} contains only
distinct drafts, including the initial draft. Fixed points take priority.
These categories describe behavior within the observed revision budget,
not guaranteed long-run dynamics.

\textbf{Repetition often persists, but its form varies across tasks and models.}
Terminal fixed points or cycles account for 69.0\%, 57.6\%, and 79.5\%
of failures in exact-length, lexical, and compositional tasks, respectively.
Fixed points alone account for 55.4\%, 48.9\%, and 71.2\%: compositional
failures are especially dominated by fixed points, whereas lexical failures
contain larger shares of other repetition and no repetition.
Controller differences also matter. For example, on exact-length tasks, Qwen3 1.7B's failures
are predominantly fixed points, whereas Granite and Falcon-H1 have large
nonrepeating shares despite their low final success.
\textbf{These rates are conditional on failure.} In particular, a large
repeated-output share among a strong model's few failures does not imply
frequent repetition overall. Recurrence is an important failure pattern,
not a complete explanation of model quality. Appendix~\ref{app:all-constraint-failures}
reports the counts and formal classification rules.

\textbf{Prior recurrence is associated with lower subsequent recoverability.}
For exact-length tasks, we restrict the analysis to trajectories that remain unresolved after revision 4 and have at least one subsequent revision. We then test whether recurrence observed through revision 4 is associated with successful recovery during revisions 5--8. To distinguish this association from differences in the current revision state, we adjust for current error and missing anchors, prior error reduction and editing behavior, checkpoint, source, and length band (full specification in Appendix~\ref{app:controlled-interventions}).

\begin{wraptable}[12]{r}{0.50\textwidth}
\centering\scriptsize
\setlength{\tabcolsep}{2pt}
\begin{tabular*}{\linewidth}{@{\extracolsep{\fill}}lcccc@{}}
\toprule
\textbf{Panel} & \textbf{Size} $n$ & \textbf{Odds Ratio}& 95\% \textbf{CI} & $p$ \textbf{Value} \\
\midrule
Local & 5,039 & 0.059 & 0.019--0.186 & 1.19$\!\times\!$10$^{\text -6}$ \\
API & 1,235 & 0.017 & 0.002--0.120 & 4.70$\!\times\!$10$^{\text -5}$ \\
All & 6,274 & 0.047 & 0.017--0.130 & 3.60$\!\times\!$10$^{\text -9}$ \\
\bottomrule
\end{tabular*}
\caption{Adjusted odds of success in revisions 5--8 among cases unresolved at revision 4. Odds ratio compares repeat fractions 1 versus 0; $n$ is the risk-set size. Intervals and $p$ values cluster by case.}
\label{tab:recurrence-rescue-or}
\end{wraptable}

Table~\ref{tab:recurrence-rescue-or} shows a consistent negative association between prior recurrence and later recovery. An odds ratio below 1 indicates that trajectories with more prior repetition have lower odds of succeeding during revisions 5--8 after accounting for the measured revision state and history. In the local panel, increasing the prior repeat fraction from 0 to 1 corresponds to an adjusted odds ratio of 0.059, meaning that the estimated odds of later success fall to about 5.9\% of those at the no-repeat end of the scale. The API panel shows the same pattern (OR $=0.017$), with a combined estimate of 0.047. All three 95\% confidence intervals lie well below the no-association value of 1, with small $p$ values, supporting a robust negative association. The same direction is observed across most individual controllers and data sources, with additional landmark and length-subset checks reported in Appendix~\ref{app:controlled-interventions}.

Recurrence therefore carries information about later recoverability even after adjustment for the measured state and history. This is a predictive association, not evidence that repetition itself causes failure. To test whether retained dialogue contributes to continued recurrence, we intervene on earlier history while holding the current draft and verifier feedback fixed, and measure subsequent revision behavior and final success separately in the next subsection.

\subsection{History Interventions at Matched States}
\label{sec:history-reset}
To move from association to intervention, we compare continuations from the same recurrent state while varying only the retained dialogue history. The full-history condition preserves earlier draft--feedback pairs, whereas the reset condition removes them. The task, current draft, verifier report, decoding settings, and remaining revision budget are otherwise identical, allowing us to isolate how retained history changes subsequent revision. Our primary exact-length analysis includes 83 Llama 3.1 8B and 120 GLM-4 9B recurrent states selected at revision three or later, with over-length Llama drafts and under-length GLM drafts (Appendix~\ref{app:history-reset}).

\begin{wrapfigure}{r}{0.48\textwidth}
\centering
\setlength{\abovecaptionskip}{2pt}
\includegraphics[width=\linewidth]{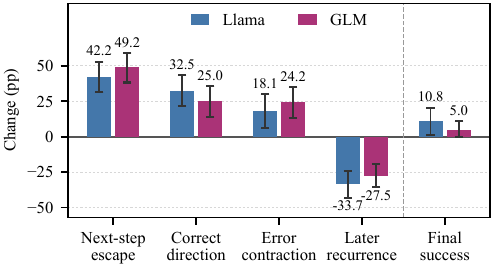}
\caption{Reset-minus-full effects on selected exact-length states; stratified paired-bootstrap 95\% intervals. Lower recurrence is better.}
\label{fig:interventions-combined}
\end{wrapfigure}

\begin{figure}[!t]
\centering
\includegraphics[width=\textwidth]{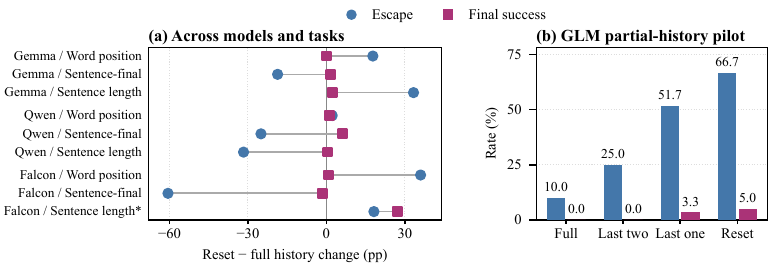}
\vspace{-4pt}
\setlength{\abovecaptionskip}{3pt}
\caption{History effects beyond the primary exact-length intervention. (a) Reset-minus-full changes for nine model--task cells (868 states); lines pair outcomes. *Descriptive cell ($n=11$). (b) Four categorical history conditions on the same 60 GLM states, task-balanced; last two/one retain that many draft--feedback pairs.}
\label{fig:history-beyond-primary}
\end{figure}

\textbf{Removing earlier history substantially changes continuation behavior, but much less reliably improves final success.}
As Figure~\ref{fig:interventions-combined} shows, next-step escape increases by 42.2 percentage points for Llama and 49.2 for GLM. Correct-direction editing increases by 32.5 and 25.0 points, error contraction by 18.1 and 24.2 points, and later recurrence falls by 33.7 and 27.5 points. These changes go beyond merely producing a different output: after reset, continuations are more likely to move toward the target, reduce word-count error, and avoid returning to previously visited outputs. Yet the gains in final success are much smaller, increasing by only 10.8 points for Llama and 5.0 for GLM, to 21.7\% and 8.3\%, respectively. Most reset continuations therefore still fail to satisfy all constraints within the remaining budget. \textbf{History removal can strongly reshape local revision behavior without producing proportionate improvements in closed-loop completion.} Figure~\ref{fig:history-beyond-primary} broadens this result beyond the primary exact-length intervention.

\par\WFclear

\textbf{History effects vary across models and tasks, but local escape remains weakly coupled to completion.}
Across 868 matched recurrent states from Gemma, Qwen, and Falcon on lexical and compositional tasks, removing earlier dialogue changes escape from $-60.6$ to $+36.0$ percentage points across the nine model--task cells (Figure~\ref{fig:history-beyond-primary}a). Final-success changes are much smaller in eight of the nine cells, ranging from $-1.4$ to $+6.2$ points; the remaining Falcon sentence-length cell contains only 11 states and is reported descriptively. A 60-state GLM partial-history pilot shows the same separation: retaining progressively less dialogue raises escape from 10.0\% under full history to 66.7\% after reset, while final success remains between 0.0\% and 5.0\% (Figure~\ref{fig:history-beyond-primary}b). These results show that history removal can help or hinder recurrence escape depending on the model and task, while large local changes need not translate into comparable gains in completion. The direction also depends on the trigger-state composition: excluding first-revision repeats reverses some escape effects (Appendix~\ref{app:trigger-sensitivity}).

Taken together, the matched-state experiments show that retained dialogue is part of the effective revision state: changing history can substantially alter continuation behavior even when the current draft and feedback are fixed. \textbf{History therefore shapes closed-loop dynamics, but manipulating history alone is not sufficient to make revision reliable.} Full cross-task results, paired partial-history contrasts, and trigger-state sensitivity are reported in Appendices~\ref{app:crossconstraint-history}, \ref{app:partial-history}, and~\ref{app:trigger-sensitivity}. Appendix~\ref{app:localization} reports supplementary internal-model analyses.

\section{Discussion and Conclusion}
\label{sec:discussion}

Taken together, our results suggest that closed-loop revision should be understood as a property of the interaction between the model and its evolving revision state, rather than of feedback quality alone. Once feedback correctness and completeness are held fixed, substantial differences remain across models and constraints, persist from matched starting drafts, and appear as reproducible feedback-to-action response patterns. These differences do not follow a simple ordering by model scale or deployment category. At the trajectory level, recurrent outputs mark states with lower subsequent recoverability, while matched-history interventions show that changing retained dialogue can strongly alter local continuation behavior without reliably improving final completion. The same intervention can even help or hinder recurrence escape across models and tasks. Together, these findings separate three notions that are often conflated in iterative revision: observing the error, translating feedback into an effective action, and reliably completing the loop.

Evaluation should therefore consider not only endpoint success or single-step improvement, but also consistent feedback incorporation, recurrence, and retained history. More broadly, improvements at one stage need not propagate to end-to-end completion: changing history can substantially alter local revision behavior with much smaller gains in final success. Our fixed-budget, deterministically checkable text tasks do not cover subjective objectives, incomplete feedback, or changing environments. Within this scope, \textbf{exact feedback provides observability, not control: reliable closed-loop revision requires models whose responses remain effective as the revision state evolves.}

\subsection*{Reproducibility statement}
Section~\ref{sec:assay} and Appendix~\ref{app:reproducibility} specify verifiers,
data, controllers, decoding, interventions, and provenance limits. The public
repository provides experiment and analysis code, configurations, necessary
frozen inputs, tests, and reproduction instructions. Frozen experimental
outputs and the evidence package are not included; offline reanalysis requires
separately supplied evidence. Reruns require user-supplied checkpoints or API
credentials; neither is included.

\subsection*{AI use statement}

Generative AI tools were used to assist with methodological and experimental-design
feedback, implementation, data processing and analysis.
They were also used for literature search and summarization, figure and artifact
preparation, and manuscript drafting and editing. Generative AI was not used to
generate the evaluation datasets or to formulate or prove mathematical claims. All
AI-assisted research outputs were reviewed and verified by the authors, who
conducted and audited the reported experiments and take full responsibility
for the final content of this work.

\bibliographystyle{iclr2027_conference}
\bibliography{refs}

\clearpage
\appendix
\renewcommand{\topfraction}{0.9}
\renewcommand{\bottomfraction}{0.8}
\renewcommand{\textfraction}{0.08}
\renewcommand{\floatpagefraction}{0.85}
\setcounter{topnumber}{4}
\setcounter{bottomnumber}{3}
\setcounter{totalnumber}{6}
\setlength{\textfloatsep}{9pt plus 2pt minus 2pt}
\setlength{\floatsep}{8pt plus 2pt minus 2pt}
\setlength{\intextsep}{8pt plus 2pt minus 2pt}
\makeatletter
\setlength{\@fptop}{0pt}
\setlength{\@fpsep}{10pt}
\setlength{\@fpbot}{0pt plus 1fil}
\makeatother
\AtBeginEnvironment{table}{\footnotesize\setlength{\tabcolsep}{4pt}\renewcommand{\arraystretch}{1.0}\setlength{\abovecaptionskip}{4pt}}

\section{Constraint Families, Data, and Execution}
\label{app:reproducibility}

\paragraph{Reading guide.}
Appendix A specifies the tasks, data, and budget replay; Appendix B gives
endpoint inference and robustness checks. Appendix C supports the controlled
response, recoverability, and history analyses in Sections~\ref{sec:response-curves}--\ref{sec:history-reset}.
Appendices D and E report supplementary checkpoint interventions and interface
comparisons; Appendix F covers crossover, capture dynamics, and trigger-state
sensitivity, and Appendix G gives the common failure taxonomy and counts.

\subsection{Verifiers and Violation Measures by Constraint Family}
\label{app:verifiers}

For every constraint family, the fixed verifier returns a joint-success
indicator $J$ and a nonnegative violation vector $\mathbf{v}_k$.
When an analysis requires a scalar violation measure, we use
\begin{equation}
  E_k(\mathbf{v})=\sum_j v_j,
  \qquad E_k(\mathbf{v})=0\ \Longleftrightarrow\ J=1.
  \label{eq:violation-energy}
\end{equation}
This sum summarizes violations within a task; its components retain
their distinct meanings and units.

\paragraph{Exact length.}
The frozen Unicode-aware word count $W(y)$ uses the regular expression
\begin{center}
\texttt{\textbackslash b[\textbackslash w]+(?:[-'][\textbackslash w]+)*\textbackslash b}.
\end{center}
Each case has target $N$ and anchor set $A$, extracted before generation from
the human-authored prompt--reference pair. The matcher $M(q,y)$ is
case-insensitive, token-boundary-aware, and permits flexible whitespace. The
reported endpoint is
\begin{equation}
  J_{\min}(y)=\mathbf{1}[W(y)=N\ \wedge\ \forall q\in A:M(q,y)=1].
  \label{eq:jmin}
\end{equation}
For this constraint family, $J_t=J_{\min}(y_t)$.

Its violation vector contains normalized absolute length error followed by one
missingness indicator per anchor,
\begin{equation}
  \mathbf{v}_{\mathrm{len}}(y)=
  \left(\frac{|W(y)-N|}{\max(N,1)},
  \{1-M(q,y):q\in A\}\right).
  \label{eq:length-energy}
\end{equation}
Any stricter binary content requirement $Q$ can only reduce the accepted set:
$J_{\min}(y)Q(y)\leq J_{\min}(y)\leq\mathbf{1}[W(y)=N]$. Anchors establish
minimum retention; semantic equivalence and subjective quality remain
unmeasured. Energy analyses across constraint families use Equation~\ref{eq:length-energy};
signed system-identification and exact-length reset analyses explicitly use
$|d_t|$ when reporting absolute length-error contraction.

\paragraph{Lexical and compositional structures.}
Let $n_w,n_s$ be the observed word and sentence counts and $N_w,N_s$
their requested values. The count residuals are $r_w=n_w-N_w$ and
$r_s=n_s-N_s$. Let $\delta_j$ indicate a missing or mismatched lexical condition,
$f_j$ count occurrences of forbidden word $j$, and $n_i$ be the word count of
observed sentence $i$. The five COLLIE-derived structures use
\begin{align}
\mathbf{v}_{\mathrm{c05}}
  &=\left(|r_w|/\max(N_w,1),\delta^{\mathrm{pos}}_1,\ldots,
  \delta^{\mathrm{pos}}_q\right),\\
\mathbf{v}_{\mathrm{c07}}
  &=\left(\delta^{\mathrm{req}}_1,\ldots,\delta^{\mathrm{req}}_q\right),\\
\mathbf{v}_{\mathrm{c09}}
  &=\left(|r_s|/\max(N_s,1),f_1,\ldots,f_q\right),\\
\mathbf{v}_{\mathrm{c10}}
  &=\left(|r_s|/\max(N_s,1),
  \{(L-n_i)_+/\max(L,1)\}_{i=1}^{n_s},
  \{(n_i-U)_+/\max(U,1)\}_{i=1}^{n_s}\right),\\
\mathbf{v}_{\mathrm{c12}}
  &=\left(|r_s|/\max(N_s,1),\delta^{\mathrm{final}}_1,\ldots,
  \delta^{\mathrm{final}}_q\right).
\label{eq:structured-violation-vectors}
\end{align}
Here $q$ is the number of lexical conditions in the task, $L$ and $U$
are the requested per-sentence bounds, and $(z)_+=\max(z,0)$.
In words, c05 imposes word-position requirements under a word-count target,
c07 required-word presence, c09 forbidden words under a sentence-count target,
c10 per-sentence length bounds under a sentence-count target, and c12
sentence-final words under a sentence-count target. The lexical constraint family allocates
240/120/120 of its 480 cases to c05/c07/c12 and the compositional evaluation set
240/240 to c09/c10. Thus the lexical column of
Table~\ref{tab:all-model-assay-matrix} is an unequally weighted composite;
the compositional column equally weights its two structures.
Because c05 and c12 also carry count residuals, the three family labels are
reporting conventions rather than a partition by violation type.
Frozen word tokenization and sentence splitting are shared by generation and
audit; each case includes a human-authored witness that passes the same
verifier.

\paragraph{Trajectory events.}
An unsuccessful recurrence occurs when the SHA-256 hash of a draft equals that
of an earlier draft while $J=0$. An unresolved trajectory is a persistent fixed
point when its final three drafts are identical. It is a terminal period-$p$
cycle, $p\in\{2,3,4\}$, when its final $2p$ drafts comprise two identical
$p$-draft blocks and the block contains at least two distinct drafts. For a
selected recurrent state, first-step escape means that the next draft differs
from every draft seen before the intervention. Later recurrence checks for
repetition of any previously seen draft after the first continuation step;
final joint success is measured separately.

\paragraph{Exact-length failure categories.}
\label{app:failure-categories}
The additional exact-length classification covers protocol-complete
trajectories without final joint success: 4,622 local and 650
API failures. Protocol-incomplete records are excluded. For a trajectory
with drafts $y_0,\ldots,y_K$, let $K$ be its last recorded revision index
and $d_t=W(y_t)-N$ its signed word-count error. All checks include the
initial draft. The classifier assigns the \emph{first matching category}
in the following order; the legend order is only for display.
\begin{enumerate}
\item \textbf{Anchor failure.} The final draft has exactly the target word
count ($d_K=0$) but is missing at least one required anchor. This condition
is checked at the endpoint and takes priority over all trajectory patterns.
\item \textbf{Recurrence.} Two recorded drafts are byte-identical:
$y_i=y_j$ for some $0\leq i<j\leq K$, checked by SHA-256 of their UTF-8
text without normalization. The repeat need not be consecutive or terminal.
This category is therefore broader than a persistent fixed point or terminal
cycle, as defined above.
\item \textbf{Sign flips ($\geq2$).} After removing zero errors from
$(d_0,\ldots,d_K)$, the remaining sign sequence changes at least twice.
For example, errors $(3,0,-2,1)$ give two flips. These are crossings of the
word-count target, not necessarily repeated outputs or growing errors.
\item \textbf{Repeated residual.} At least one consecutive pair has
$d_{t+1}=d_t$. The word count is unchanged at that step, although the text
may change. Byte-identical outputs receive the earlier recurrence label.
\item \textbf{Near target ($\leq2$).} The best absolute error anywhere in
the trajectory satisfies $\min_{0\leq t\leq K}|d_t|\leq2$. This includes
zero error and does not require the final draft to remain near the target.
Exact word count alone does not establish joint success.
\item \textbf{Net contraction.} The final absolute error is smaller than
the initial one: $|d_K|<|d_0|$. Intermediate errors may increase, so this
label does not imply monotonic improvement or that a larger budget would
suffice for success.
\item \textbf{Other non-contraction.} None of the preceding conditions
holds. In particular, $|d_K|\geq|d_0|$; equality here compares the endpoints
and need not imply equal errors at consecutive revisions.
\end{enumerate}
The assigned labels partition failures, but their underlying conditions can
overlap. For example, a trajectory may both repeat a draft and finish closer
to the target; it is labeled recurrence. Category shares thus describe this
fixed classification rule, not the prevalence of independent causal
mechanisms. The common output-repetition classification used in
Figure~\ref{fig:failure-anatomy} is defined separately in
Appendix~\ref{app:all-constraint-failures}.

\subsection{Data Construction and Frozen Execution}

Combined-480 contains 120 cases each from Dolly
\citep{databricks2023dolly15k}, No Robots
\citep{huggingfaceh42023norobots}, WritingPrompts
\citep{fan-etal-2018-hierarchical}, and OASST1
\citep{kopf2023openassistant}. Source partitions and case IDs are fixed across
checkpoints. The target is the deterministic count of the human response;
literal anchors are frozen before generation. Targets span 10--500 words. The 480 cases contain
1,907 anchors, with 3--5 per case (mean 3.973; median 4); 1,804/1,907 (94.6\%)
are single-token anchors under the frozen word regex.

The lexical and compositional panels each contain 480 frozen cases constructed
from official COLLIE instances \citep{yao2023collie}. The lexical panel covers
required-word, lexical-position, and sentence-final constraints; the
compositional panel combines sentence-count requirements with forbidden words
or per-sentence length bounds. Each panel contains 160 cases from CC-News, 160
from Gutenberg, and 160 from Wikipedia. COLLIE supplies the prompts,
constraint targets, and human-authored feasible witnesses; we select, balance,
audit, and freeze the cases before evaluation. Case IDs are shared across
controllers within a constraint family and are distinct across families.

The task, target fields, prompt format, revision budget, and stopping
rule are fixed before observing outcomes. Round zero supplies the task and response-only instruction.
After each failure, the model receives its complete previous response and the
full deterministic verifier report. The loop stops at joint success or after
eight revisions.

\subsection{Controller Roster and Main-Panel Inference}

The 12 locally executed checkpoints are Llama 3.1 8B
\citep{grattafiori2024llama}, Gemma 2 9B \citep{gemmateam2024gemma2}, GLM-4 9B
\citep{glmteam2024chatglm4}, Ministral 8B
\citep{mistralai2024ministral8b}, Qwen3 1.7B, 4B, 8B, 14B, 32B, and 30B-A3B
\citep{yang2025qwen}, Granite 3.3 8B \citep{ibmgranite2025granite33}, and
Falcon-H1 7B \citep{zuo2025falconh1}. They span seven model families and
include dense, MoE, and Transformer--SSM hybrid architectures. The seven API
controllers are GPT-5.6 Sol \citep{openai2026gpt56}, Claude Opus 5
\citep{anthropic2026opus5}, Gemini 3.6 Flash
\citep{google2026gemini36flash}, GLM-5.2 \citep{zhipu2026glm52}, DeepSeek V4
Flash \citep{deepseek2026v4flash}, Qwen3.7 Plus
\citep{alibaba2026qwen37plus}, and Llama 3.1 70B
\citep{grattafiori2024llama}.

For Qwen3.7 Plus, the saved run configurations across all three constraint
families specify the OpenRouter request identifier
\texttt{qwen/qwen3.7-plus}. The accompanying catalog snapshots record
\texttt{qwen/qwen3.7-plus-20260602} as the canonical slug. This catalog
identifier does not establish the backend weight version for each request;
we do not infer a historical snapshot from the provider's current alias mapping.

Evaluating every controller on every constraint family yields 27,360 attempted trajectories.
One Gemini refusal leaves 3,359 protocol-complete API exact-length records
and 479 cases common to all seven API controllers for paired inference.
Other combinations of controllers and constraint families have 480 protocol-complete cases.
Uncertainty intervals use 20,000 source-stratified paired-case bootstrap draws
\citep{efron1993bootstrap}; paired binary contrasts use exact McNemar tests
\citep{mcnemar1947sampling} with Holm correction
\citep{holm1979sequentially}; and interaction models cluster by case
\citep{cameron2015practitioners}.

\subsection{Revision-Budget Sensitivity}
\label{app:budget-sensitivity}

\begingroup
\raggedbottom
\setlength{\intextsep}{7pt plus 1pt minus 1pt}

The primary endpoint compares models after at most eight revisions,
using a budget fixed before the main experiments. We replay the completed trajectories offline under every observed revision cap
$T\in\{0,1,\ldots,8\}$. The API
exact-length replay uses the same 479 all-controller-common cases as the paired
analysis; all other combinations of controllers and constraint families use 480 cases.

\begin{table}[H]
\centering
\small
\setlength{\tabcolsep}{5pt}
\begin{tabular}{@{}ccccc@{}}
\toprule
\textbf{Revision Cap} $T$ & \textbf{Median Cell Success} & \textbf{Cell Range} & \textbf{Local Reversals} & \textbf{API Reversals} \\
\midrule
0 & 26.0\% & 0.2--98.5\% & 28/66 & 4/21 \\
1 & 31.9\% & 0.6--99.8\% & 21/66 & 7/21 \\
2 & 35.8\% & 1.0--100.0\% & 18/66 & 8/21 \\
3 & 40.6\% & 1.2--100.0\% & 23/66 & 7/21 \\
4 & 42.5\% & 1.2--100.0\% & 20/66 & 7/21 \\
5 & 51.6\% & 1.2--100.0\% & 19/66 & 8/21 \\
6 & 54.8\% & 1.2--100.0\% & 18/66 & 8/21 \\
7 & 54.8\% & 1.5--100.0\% & 17/66 & 8/21 \\
8 & 54.8\% & 1.5--100.0\% & 17/66 & 8/21 \\
\bottomrule
\end{tabular}
\caption{Frozen revision-budget replay across all 57 combinations of controllers and constraint families.
A reversing pair has opposite strict controller orderings on at least two
constraint families at the same cap; ties do not count as strict reversals.}
\label{tab:budget-sensitivity}
\end{table}

Absolute success changes with the action budget. Both panels retain multiple
cross-family ordering reversals at every inspected cap. Their existence
is therefore robust to the budget cap, although individual rankings change.

\paragraph{Extended-budget audit.}
The revision-32 audit selects 144 stratified revision-8 failures. Twelve succeed
by revision 32, 113 remain unsuccessful through revision 32, and 19 exhaust the native context
window. Equivalently, 12 of the 125 context-feasible cases succeed. Llama
succeeds on 10/24 additional cases, Gemma and Qwen3 14B on 1/24
each, and GLM, Ministral, and Qwen3 8B on none; 18/24 Gemma trajectories
and 1/24 Qwen3 8B trajectories terminate at the context limit.
This selected-failure cohort does not estimate population-wide gains from a longer budget.

Tables~\ref{tab:budget-full-exact-length}--\ref{tab:budget-full-compositional}
report cumulative success for every model and constraint family at each
revision cap, showing where success rates plateau and where they continue
to rise late in the budget.

\begin{table}[H]
\centering
\footnotesize
\setlength{\tabcolsep}{3.5pt}
\renewcommand{\arraystretch}{1.00}
\begin{tabular*}{\textwidth}{@{\extracolsep{\fill}}lrrrrrrrrr@{}}
\toprule
\textbf{Controller} & \textbf{0} & \textbf{1} & \textbf{2} & \textbf{3} & \textbf{4} & \textbf{5} & \textbf{6} & \textbf{7} & \textbf{8} \\
\midrule
\multicolumn{10}{@{}l}{\emph{Local controllers}} \\
Llama 3.1 8B & 3.5 & 12.5 & 25.8 & 41.3 & 56.5 & 67.3 & 74.6 & 77.3 & 79.4 \\
Qwen3 32B & 0.8 & 3.1 & 6.0 & 10.8 & 16.9 & 22.3 & 25.0 & 27.5 & 30.0 \\
Gemma 2 9B & 0.4 & 3.1 & 7.3 & 12.1 & 16.5 & 20.6 & 24.4 & 28.3 & 29.6 \\
Falcon-H1 7B & 0.8 & 1.5 & 3.1 & 6.5 & 9.0 & 12.1 & 15.8 & 17.9 & 20.4 \\
Qwen3 14B & 2.3 & 5.2 & 9.4 & 12.7 & 14.0 & 14.4 & 15.2 & 15.6 & 15.6 \\
Qwen3 30B-A3B & 1.9 & 2.3 & 3.3 & 5.8 & 7.3 & 10.2 & 11.5 & 13.3 & 15.4 \\
Qwen3 8B & 1.9 & 3.1 & 4.4 & 5.6 & 7.1 & 7.9 & 9.0 & 9.0 & 9.6 \\
GLM-4 9B & 0.2 & 0.8 & 1.9 & 2.3 & 3.5 & 5.0 & 6.0 & 8.3 & 10.4 \\
Qwen3 4B & 1.3 & 3.5 & 4.6 & 6.0 & 6.3 & 6.7 & 6.9 & 7.1 & 7.1 \\
Granite 3.3 8B & 1.0 & 2.1 & 4.2 & 6.3 & 7.9 & 9.0 & 10.2 & 11.3 & 12.3 \\
Ministral 8B & 0.2 & 0.8 & 1.7 & 3.3 & 4.2 & 5.0 & 5.2 & 5.6 & 5.8 \\
Qwen3 1.7B & 0.2 & 0.6 & 1.0 & 1.3 & 1.3 & 1.3 & 1.3 & 1.5 & 1.5 \\
\midrule
\multicolumn{10}{@{}l}{\emph{API controllers}} \\
GPT-5.6 Sol & 5.2 & 32.4 & 63.0 & 84.3 & 96.0 & 97.9 & 99.4 & 99.6 & 99.6 \\
Claude Opus 5 & 7.3 & 42.4 & 71.8 & 84.3 & 90.6 & 95.6 & 97.3 & 98.3 & 99.0 \\
Gemini 3.6 Flash & 6.7 & 25.1 & 48.0 & 63.9 & 77.0 & 85.4 & 90.0 & 94.2 & 96.2 \\
GLM-5.2 & 3.8 & 9.0 & 19.8 & 30.5 & 42.2 & 51.6 & 58.0 & 64.7 & 68.5 \\
DeepSeek V4 Flash & 0.8 & 6.3 & 14.2 & 24.8 & 41.1 & 55.1 & 64.1 & 67.8 & 70.1 \\
Llama 3.1 70B & 2.3 & 13.4 & 34.9 & 53.9 & 69.3 & 77.5 & 83.3 & 85.4 & 88.1 \\
Qwen3.7 Plus & 2.9 & 8.4 & 13.4 & 19.8 & 26.3 & 29.9 & 35.5 & 39.9 & 43.0 \\
\bottomrule
\end{tabular*}
\caption{Exact-length cumulative joint success (\%) after each revision cap. API exact-length cells use the 479-case all-controller-common set; all other cells use 480 cases.}
\label{tab:budget-full-exact-length}
\end{table}
\begin{table}[H]
\centering
\footnotesize
\setlength{\tabcolsep}{3.5pt}
\renewcommand{\arraystretch}{1.00}
\begin{tabular*}{\textwidth}{@{\extracolsep{\fill}}lrrrrrrrrr@{}}
\toprule
\textbf{Controller} & \textbf{0} & \textbf{1} & \textbf{2} & \textbf{3} & \textbf{4} & \textbf{5} & \textbf{6} & \textbf{7} & \textbf{8} \\
\midrule
\multicolumn{10}{@{}l}{\emph{Local controllers}} \\
Llama 3.1 8B & 33.3 & 40.2 & 44.6 & 47.1 & 48.1 & 49.4 & 50.2 & 50.6 & 50.8 \\
Qwen3 32B & 24.6 & 32.7 & 35.8 & 37.1 & 38.3 & 39.0 & 39.6 & 40.0 & 40.2 \\
Gemma 2 9B & 27.9 & 31.0 & 32.9 & 33.3 & 33.5 & 34.0 & 34.0 & 34.2 & 34.2 \\
Falcon-H1 7B & 25.6 & 27.1 & 28.3 & 29.2 & 30.2 & 30.2 & 30.6 & 30.6 & 30.6 \\
Qwen3 14B & 28.3 & 35.2 & 39.8 & 40.6 & 42.5 & 42.9 & 43.1 & 43.3 & 43.3 \\
Qwen3 30B-A3B & 24.2 & 27.5 & 29.4 & 30.0 & 30.4 & 30.4 & 30.6 & 30.6 & 30.6 \\
Qwen3 8B & 26.7 & 31.5 & 32.5 & 33.5 & 34.2 & 34.6 & 34.6 & 35.2 & 35.4 \\
GLM-4 9B & 22.9 & 24.0 & 24.8 & 25.4 & 25.8 & 26.3 & 26.3 & 26.3 & 26.3 \\
Qwen3 4B & 21.7 & 26.5 & 28.8 & 29.2 & 29.4 & 29.4 & 29.8 & 29.8 & 29.8 \\
Granite 3.3 8B & 20.0 & 21.7 & 23.1 & 23.1 & 23.1 & 23.3 & 23.3 & 23.3 & 23.8 \\
Ministral 8B & 17.7 & 22.7 & 26.3 & 27.1 & 27.1 & 27.1 & 27.1 & 27.1 & 27.1 \\
Qwen3 1.7B & 22.7 & 23.8 & 24.2 & 24.2 & 24.2 & 24.2 & 24.2 & 24.2 & 24.2 \\
\midrule
\multicolumn{10}{@{}l}{\emph{API controllers}} \\
GPT-5.6 Sol & 71.9 & 91.3 & 97.3 & 98.8 & 99.4 & 99.4 & 99.6 & 99.8 & 99.8 \\
Claude Opus 5 & 52.9 & 64.8 & 74.0 & 84.6 & 92.3 & 95.8 & 96.9 & 98.1 & 98.8 \\
Gemini 3.6 Flash & 51.3 & 56.0 & 60.4 & 66.0 & 70.4 & 73.5 & 76.9 & 79.4 & 82.3 \\
GLM-5.2 & 46.7 & 52.5 & 54.6 & 56.3 & 57.3 & 58.1 & 58.8 & 58.8 & 58.8 \\
DeepSeek V4 Flash & 44.8 & 50.2 & 52.5 & 53.5 & 53.8 & 54.4 & 54.8 & 54.8 & 54.8 \\
Llama 3.1 70B & 16.0 & 31.9 & 35.2 & 36.3 & 37.7 & 40.0 & 40.8 & 41.9 & 42.9 \\
Qwen3.7 Plus & 47.3 & 56.5 & 61.3 & 63.1 & 64.8 & 65.2 & 65.2 & 65.4 & 65.4 \\
\bottomrule
\end{tabular*}
\caption{Lexical cumulative joint success (\%) after each revision cap. Every cell uses 480 cases.}
\label{tab:budget-full-lexical}
\end{table}
\begin{table}[H]
\centering
\footnotesize
\setlength{\tabcolsep}{3.5pt}
\renewcommand{\arraystretch}{1.00}
\begin{tabular*}{\textwidth}{@{\extracolsep{\fill}}lrrrrrrrrr@{}}
\toprule
\textbf{Controller} & \textbf{0} & \textbf{1} & \textbf{2} & \textbf{3} & \textbf{4} & \textbf{5} & \textbf{6} & \textbf{7} & \textbf{8} \\
\midrule
\multicolumn{10}{@{}l}{\emph{Local controllers}} \\
Llama 3.1 8B & 71.3 & 88.5 & 96.3 & 96.5 & 96.9 & 97.1 & 97.3 & 97.3 & 97.3 \\
Qwen3 32B & 75.0 & 92.9 & 95.8 & 97.7 & 97.7 & 97.9 & 97.9 & 97.9 & 97.9 \\
Gemma 2 9B & 61.0 & 67.7 & 71.0 & 71.7 & 72.1 & 72.5 & 72.9 & 74.0 & 74.2 \\
Falcon-H1 7B & 36.7 & 41.7 & 69.4 & 73.3 & 76.7 & 78.1 & 80.0 & 80.6 & 80.6 \\
Qwen3 14B & 40.6 & 48.8 & 65.2 & 65.8 & 67.7 & 69.2 & 69.2 & 70.0 & 70.6 \\
Qwen3 30B-A3B & 62.1 & 71.7 & 72.7 & 72.9 & 73.3 & 73.8 & 73.8 & 74.0 & 74.0 \\
Qwen3 8B & 59.2 & 60.0 & 67.3 & 68.1 & 68.3 & 69.4 & 69.4 & 69.4 & 69.4 \\
GLM-4 9B & 48.5 & 50.8 & 56.9 & 62.3 & 62.7 & 62.7 & 62.9 & 62.9 & 62.9 \\
Qwen3 4B & 53.1 & 57.1 & 59.2 & 60.2 & 60.2 & 60.2 & 60.2 & 60.4 & 60.4 \\
Granite 3.3 8B & 31.5 & 37.7 & 39.4 & 39.8 & 40.4 & 40.4 & 40.4 & 40.4 & 40.4 \\
Ministral 8B & 28.1 & 28.8 & 28.8 & 28.8 & 28.8 & 28.8 & 28.8 & 28.8 & 28.8 \\
Qwen3 1.7B & 26.0 & 26.5 & 26.5 & 26.5 & 26.5 & 26.5 & 26.5 & 26.5 & 26.5 \\
\midrule
\multicolumn{10}{@{}l}{\emph{API controllers}} \\
GPT-5.6 Sol & 96.5 & 99.8 & 100.0 & 100.0 & 100.0 & 100.0 & 100.0 & 100.0 & 100.0 \\
Claude Opus 5 & 98.5 & 99.0 & 99.2 & 99.2 & 99.2 & 99.2 & 99.2 & 99.2 & 99.2 \\
Gemini 3.6 Flash & 96.9 & 99.2 & 99.4 & 99.8 & 99.8 & 99.8 & 99.8 & 99.8 & 99.8 \\
GLM-5.2 & 85.8 & 96.0 & 98.8 & 99.6 & 99.6 & 99.6 & 99.6 & 99.8 & 99.8 \\
DeepSeek V4 Flash & 77.1 & 90.6 & 94.0 & 95.8 & 96.5 & 97.1 & 97.1 & 97.1 & 97.1 \\
Llama 3.1 70B & 56.9 & 70.0 & 78.1 & 81.3 & 83.5 & 84.6 & 85.6 & 86.3 & 86.7 \\
Qwen3.7 Plus & 91.9 & 95.4 & 97.7 & 97.9 & 98.1 & 98.5 & 98.5 & 98.5 & 98.5 \\
\bottomrule
\end{tabular*}
\caption{Compositional cumulative joint success (\%) after each revision cap. Every cell uses 480 cases.}
\label{tab:budget-full-compositional}
\vspace{0.45em}
\begin{minipage}{\textwidth}
\small\textit{Budget profiles.} Exact length remains budget-sensitive late: from $T=4$ to $T=8$, DeepSeek V4 Flash, GLM-5.2, and Llama 3.1 8B gain 29.0, 26.3, and 22.9 percentage points, respectively. Lexical cells generally flatten earlier, although Gemini 3.6 Flash gains 11.9 points after $T=4$. Compositional cells are most front-loaded: the largest $T=4$-to-$T=8$ gain is 4.0 points (Falcon-H1 7B). The shared endpoint captures heterogeneous late-recovery profiles and leaves convergence unresolved.
\end{minipage}
\end{table}

\subsection{Evidence Panels}

\begin{table}[H]
\centering
\footnotesize
\setlength{\tabcolsep}{3.5pt}
\renewcommand{\arraystretch}{1.08}
\begin{tabular}{@{}>{\raggedright\arraybackslash}p{0.20\textwidth}
                    >{\raggedright\arraybackslash}p{0.30\textwidth}
                    >{\raggedright\arraybackslash}p{0.44\textwidth}@{}}
\toprule
\textbf{Module} & \textbf{Scope} & \textbf{Evaluation Target} \\
\midrule
Main closed loop & 19 controllers $\times$ 3 constraint families $\times$ 480 cases & Endpoint spread and failure dynamics \\
System identification & 10 checkpoints, 12 commands, 24 discovery + 48 confirmation cases & Feedback-to-action response curves \\
Held-out action prediction & 9 checkpoints; 3,014 eligible checkpoint--case trajectories & Model-specific versus shared prediction MAE \\
Conditional rescue & 19 controllers, revision-4 risk set & Recoverability given recurrence and current state \\
History reset & 83 Llama + 120 GLM recurrent states & Recurrence escape versus final success \\
Cross-task history & 868 states across 9 model--task cells & Heterogeneity in escape and success effects \\
Partial history & 60 GLM states, 4 categorical conditions & Escape and success under partial retention \\
Aligned checkpoint panel & Llama, Qwen, and Gemma Base/Instruct pairs & Error readout, restoration, activation transplantation, and grafts \\
Long horizon & 144 stratified revision-8 failures (125 context-feasible) & Persistence beyond the primary budget \\
\bottomrule
\end{tabular}
\caption{Scope and evaluation target of each evidence panel.}
\label{tab:evidence-panels}
\end{table}
\endgroup
\raggedbottom

\section{Inference and Robustness}

\subsection{Interaction between Controller and Constraint Family}
\label{app:interaction-inference}

Table~\ref{tab:all-model-assay-matrix} reports final joint success using
480 cases per cell, except API exact length, which uses 479 cases common to
all seven systems. Its final column gives the mean and sample standard
deviation across the three constraint-family rates. Group rows average
models within each family before computing these summaries. Since the
families use distinct datasets, ranking reversals describe performance
across these settings rather than a causal effect of constraint structure.

The frozen seven-controller API panel uses 479 all-model-common evaluable
exact-length cases after one stable Gemini refusal; lexical and compositional
cells use all 480 cases. Llama 3.1 70B minus Qwen3.7 Plus changes from $+45.1$
points on exact length to $-22.5$ on lexical constraints, a
difference-in-differences of $-67.6$ points (95\% CI $[-74.3,-60.9]$;
Holm-adjusted $p=4.08\times10^{-85}$). After Holm correction, 19 of 21
exact-length pairs and 20 of 21 lexical pairs differ in final success.
Compositional constraints place six controllers near the ceiling, leaving 9
of 21 pairwise separations.

The Wald test for the interaction between controller and constraint family gives $\chi^2(12)=1214.05$
($p=1.63\times10^{-252}$), and the controller-by-structure test gives
$\chi^2(16)=3662.21$ ($p<10^{-300}$). The largest $p$-values across all
leave-one-controller-out and leave-one-structure-out refits are
$2.49\times10^{-160}$ and $7.84\times10^{-11}$, respectively. These interaction
estimates compare cells across distinct datasets for the constraint families and COLLIE case sets.

\begin{table}[h]
\centering
\scriptsize
\setlength{\tabcolsep}{4.0pt}
\begin{tabular}{lrrrr}
\toprule
\textbf{Controller} & \textbf{Initial Joint} & \textbf{Early Capture} & \textbf{Early Recurrence} & \textbf{Final Joint} \\
\midrule
\multicolumn{5}{l}{\emph{Local controllers}} \\
Llama 3.1 8B & 3.5 & 16.7 & 2.4 & 79.4 \\
Qwen3 32B & 0.8 & 4.2 & 43.6 & 30.0 \\
Gemma 2 9B & 0.4 & 4.3 & 17.2 & 29.6 \\
Falcon-H1 7B & 0.8 & 2.1 & 8.1 & 20.4 \\
Qwen3 14B & 2.3 & 3.1 & 67.1 & 15.6 \\
Qwen3 30B-A3B & 1.9 & 1.4 & 65.7 & 15.4 \\
Qwen3 8B & 1.9 & 1.4 & 66.8 & 9.6 \\
GLM-4 9B & 0.2 & 0.8 & 15.5 & 10.4 \\
Qwen3 4B & 1.3 & 1.3 & 59.5 & 7.1 \\
Granite 3.3 8B & 1.0 & 1.8 & 6.8 & 12.3 \\
Ministral 8B & 0.2 & 1.0 & 49.1 & 5.8 \\
Qwen3 1.7B & 0.2 & 0.3 & 70.9 & 1.5 \\
\midrule
\multicolumn{5}{l}{\emph{API controllers}} \\
GPT-5.6 Sol & 5.2 & 42.3 & 0.9 & 99.6 \\
Claude Opus 5 & 7.3 & 42.9 & 1.6 & 99.0 \\
Gemini 3.6 Flash & 6.7 & 27.4 & 0.2 & 96.2 \\
GLM-5.2 & 3.8 & 11.4 & 24.9 & 68.5 \\
DeepSeek V4 Flash & 0.8 & 11.4 & 8.6 & 70.2 \\
Llama 3.1 70B & 2.3 & 22.7 & 4.1 & 88.1 \\
Qwen3.7 Plus & 2.9 & 6.6 & 8.2 & 42.9 \\
\bottomrule
\end{tabular}
\caption{Canonical 19-controller exact-length panel. Initial/final are case-level $J_{\min}$
percentages; early events are per still-active transition into revisions 1--4.
API entries use protocol-complete records (479 for Gemini, 480 for each other system);
Table~\ref{tab:all-model-assay-matrix} and the interaction between controller and constraint family use the
479 all-model-common evaluable cases.}
\label{tab:all-controllers}
\end{table}

\begin{table}[t]
  \centering
  \small
  \setlength{\tabcolsep}{6pt}
  \begin{tabular}{@{}lrrrr@{}}
    \toprule
    \textbf{Controller}& \textbf{Exact} & \textbf{Lexical} & \textbf{Comp.} & \textbf{Mean $\pm$ SD} \\
    \midrule
    Local controllers ($n=12$) & 19.8 & 33.0 & 65.2 & $39.3\pm23.4$ \\
    API controllers ($n=7$)    & 80.6 & 71.8 & 97.3 & $83.3\pm12.9$ \\
    All controllers ($n=19$)   & 42.2 & 47.3 & 77.1 & $55.5\pm18.8$ \\
    \midrule
    Range across controllers & 1.5--99.6 & 23.8--99.8 & 26.5--100.0 & \\
    \bottomrule
  \end{tabular}
  \caption{Revision-8 joint success (\%) by panel and constraint family. Mean and SD summarize the three constraint-family columns.}
  \label{tab:assay-summary}
\end{table}

Table~\ref{tab:assay-summary} averages controllers within each panel and
then summarizes the three constraint-family means. The lexical column pools structures
at 240/120/120; composition pools its two structures at 240/240.

\subsection{Content Contract Robustness Check (Descriptive)}
\label{app:content-contract-robustness}

We apply seven additional conditions separately to the existing exact-length
API endpoint annotations. This post-hoc sensitivity check remains outside the
registered inference family and reports each condition separately. Missing
annotation fields are treated as not applicable and do not fail the added
condition; the denominator remains 480 cases per controller.

\begin{table}[h]
\centering
\footnotesize
\setlength{\tabcolsep}{3.0pt}
\begin{tabular}{p{0.28\textwidth}p{0.44\textwidth}rr}
\toprule
\textbf{Observable} & \textbf{Stored Audit Field / Check} & \shortstack{\textbf{Llama 3.1}\\\textbf{70B}} & \shortstack{\textbf{Qwen3.7}\\\textbf{Plus}} \\
\midrule
Complete entity retention & \path{draft_entity_retention} $=1$ & 85.2 & 31.3 \\
Complete proper-name retention & \path{draft_proper_name_retention} $=1$ & 87.1 & 42.5 \\
Complete symbolic-unit retention & \path{draft_symbolic_retention} $=1$ & 87.7 & 42.9 \\
No parser-fragment increase & \path{fragment_rate_change} $\leq0$ & 85.0 & 41.7 \\
No parser-agreement decrease & \path{agreement_rate_change} $\geq0$ & 80.4 & 42.3 \\
No parser-dangling increase & \path{dangling_rate_change} $\leq0$ & 87.9 & 42.5 \\
No NLI-contradiction increase & \path{adjacent_contradiction_mean_change} $\leq0$ & 42.1 & 19.4 \\
\bottomrule
\end{tabular}
\caption{Post-hoc descriptive check over single stricter content requirements.
Values are joint success rates (\%) for Llama 3.1 70B and Qwen3.7 Plus after adding each requirement to $J_{\min}$, over 480 cases per controller.}
\label{tab:content-contract-robustness}
\end{table}

Llama remains ahead under every measured condition, including a 54.0-point
gap under full applicable-entity retention. This pattern is inconsistent with
an advantage sustained solely by the minimal anchor contract. These observables
measure content retention; semantic equivalence and subjective quality remain
unmeasured.

\subsection{Inference and Execution Provenance}

Local runs use greedy decoding
(\texttt{do\_sample=false}), bfloat16, an EOS padding token, and
\texttt{max\_new\_tokens=max(128, 3N)}. Each checkpoint uses its own tokenizer
and documented chat template with a generation prompt; thinking mode is
disabled when the interface supports that option. The six-model original
Combined-480 panel has command-level provenance for every one of its 18
model--split cells. Primary-120 and Replication-120 use the serial runner;
Extension-240 uses a batch-one equivalent, with Qwen3 14B executed as two
ordered 120-case shards. A guarded batch-four Llama trial failed exact serial
equivalence on seven of eight gate cases, was aborted, and is excluded from the
canonical output. The batch-one gate matched every case, round, and generated
text.

Immutable analysis manifests record each input path, row count, and SHA-256
digest. The Qwen3 vLLM scale
panel uses one frozen 480-case interface at 1.7B, 4B, 8B, 14B, 32B, and
30B-A3B; an independent audit verifies 480 unique case IDs for all twelve
primary checkpoints. The environment record contains Python 3.10.19, PyTorch
2.9.1, and Transformers 4.57.6. Tokenizers 0.22.2 and Accelerate 1.12.0 were
recovered post hoc. Launch records omit driver versions, full weight hashes,
and a single immutable Qwen3 14B revision; the manifest marks these fields as
missing. Exact
reproduction of the Ministral row preserves the historical
\texttt{fix\_mistral\_regex=false} tokenizer behavior.

\subsection{Prompt and Decoding Robustness}
\label{app:robustness}

The same 60 case IDs were evaluated for Llama 3.1 8B, Gemma 2 9B, and
Qwen3 14B under three feedback prompt forms and three sampled decoding seeds.
Llama ranks above Gemma, which ranks above Qwen3 14B, in every condition
(Table~\ref{tab:prompt-decoding-robustness}). Absolute rates vary with the
interface; controller ordering is stable.

\begin{table}[h]
\centering
\footnotesize
\setlength{\tabcolsep}{3.2pt}
\begin{tabular}{lrrrrr}
\toprule
\textbf{Checkpoint} & \textbf{Baseline} & \textbf{Structured} & \textbf{Concise} & \textbf{Sample Mean} & \textbf{Sample Range} \\
\midrule
Llama 3.1 8B & 83.3 & 65.0 & 76.7 & 83.9 & 81.7--86.7 \\
Gemma 2 9B & 36.7 & 25.0 & 36.7 & 38.9 & 33.3--46.7 \\
Qwen3 14B & 18.3 & 11.7 & 10.0 & 13.3 & 11.7--15.0 \\
\bottomrule
\end{tabular}
\caption{Final $J_{\min}$ (\%) on the matched 60-case robustness panel.
Sample statistics summarize three decoding seeds.}
\label{tab:prompt-decoding-robustness}
\end{table}

\section{Closed-Loop and History Interventions}

\subsection{Response Estimation, Prediction, and Recoverability}
\label{app:controlled-interventions}

\paragraph{Controlled response estimation.}
The controlled response experiment holds source text fixed and varies only the
synthetic target and verifier message over requests
$\{\pm1,\pm2,\pm5,\pm10,\pm20,\pm50\}$. Unlike naturally occurring revision
states, this design exposes every checkpoint to the same drafts and correction
requests. The 24 discovery and 48 confirmation texts are disjoint.
All ten checkpoints use the same grid and median-action estimator.
Seven infeasible case--command cells imply targets below the 10-word floor
and are excluded for every checkpoint, leaving 286 discovery and 571
confirmation actions per checkpoint (8,570 actions overall).
Qwen3 32B and 30B-A3B were collected later under this same protocol; the
analysis here integrates all ten checkpoints rather than reporting separate
cohort aggregates. Model-specific curves and the shared curve are fit only on
discovery data and then held fixed. The shared curve is the median of all ten
checkpoints' discovery actions at each request. No confirmation or natural
trajectory outcomes are used to fit either predictor. This is an expanded
analysis of held-out data, not a new prospective confirmation gate.

Behavioral rates first average actions within each text and then average
texts equally, for every checkpoint. Under this common weighting,
Qwen3 30B-A3B has 59.5\% zero word-count change, 35.0\% correct direction,
and 26.9\% error contraction on confirmation texts.
Its discovery--confirmation median-curve correlation is $\rho=1.000$, with
nine of twelve request levels tied at zero in both splits; this reflects rank
reproducibility rather than correction accuracy. The corresponding correlation
for Qwen3 32B is $\rho=0.951$. The MoE comparison describes one checkpoint and
does not isolate an architecture effect from training or computation differences.

Table~\ref{tab:base-instruct-behavior} reports the Base/Instruct comparison
separately for discovery and confirmation under this same text-balanced
weighting. For requested change $c=-d_t$ and realized edit $a_t$, correct
direction means $a_tc>0$, error expansion means $|a_t-c|>|c|$, and zero
action means $a_t=0$ (unchanged word count, not necessarily identical text).
These events are not a partition: an edit in the correct direction can
overshoot far enough to increase absolute error. On confirmation texts,
Qwen instruction tuning reduces error expansion from 47.3\% to 14.2\%
while increasing zero action from 35.9\% to 49.8\%; discovery has the same
directions. These are descriptive checkpoint contrasts under the controlled
protocol, not an isolation of training from all other checkpoint differences.
\begin{table}[!htbp]
\centering
\begin{tabular*}{\linewidth}{@{\extracolsep{\fill}}llrrr@{}}
\toprule
Checkpoint & Split & \shortstack{Correct direction\\(\%)} & \shortstack{Error expansion\\(\%)} & \shortstack{Zero action\\(\%)} \\
\midrule
Llama 3.1 8B Base & Discovery & 27.4 & 41.8 & 53.0 \\
Llama 3.1 8B Instruct & Discovery & 87.5 & 24.0 & 5.9 \\
Qwen3 8B Base & Discovery & 39.5 & 40.6 & 35.3 \\
Qwen3 8B Instruct & Discovery & 44.9 & 13.8 & 52.5 \\
\midrule
Llama 3.1 8B Base & Confirmation & 30.7 & 51.2 & 46.2 \\
Llama 3.1 8B Instruct & Confirmation & 90.1 & 21.1 & 5.8 \\
Qwen3 8B Base & Confirmation & 39.1 & 47.3 & 35.9 \\
Qwen3 8B Instruct & Confirmation & 47.4 & 14.2 & 49.8 \\
\bottomrule
\end{tabular*}
\caption{Controlled Base/Instruct behavior. Discovery uses 24 texts and 286 feasible text--request pairs per checkpoint; confirmation uses 48 disjoint texts and 571 pairs. Rates average requests within each text, then weight texts equally.}
\label{tab:base-instruct-behavior}
\end{table}

\paragraph{Held-out action prediction.}
The independent natural-trajectory evaluation includes nine checkpoints;
Qwen3 8B Base has no matched Combined-480 trajectory panel. For each included
checkpoint, all 72 controlled-study texts are excluded from its 480 unique
cases, leaving 408 cases before transition filtering. We score only transitions
with $0<|c|\leq50$ words. The final sample contains 3,014 checkpoint--case
trajectories with at least one eligible transition and 20,069 transitions;
the same source case may contribute to multiple checkpoints.
Both predictors use piecewise-linear interpolation between measured requests,
with $(0,0)$ added as a construction convention rather than an observed response.

For each predictor, we average absolute prediction errors across eligible
transitions within a case, across cases within a checkpoint, and then equally
across the nine checkpoints. Let $\overline{\mathrm{MAE}}$ denote this last mean.
The single primary aggregate compares the model-specific and shared predictors:
\[
R = \frac{\overline{\mathrm{MAE}}_{\mathrm{shared}}
         -\overline{\mathrm{MAE}}_{\mathrm{specific}}}
        {\overline{\mathrm{MAE}}_{\mathrm{shared}}}.
\]
The resulting MAEs are 17.45 and 15.54 words, respectively: a reduction of
1.91 words (95\% CI 1.58--2.25), or 11.0\% (95\% CI 8.5--13.9).
Relative reduction is computed after averaging MAEs, not by averaging
checkpoint-level percentages. All prediction intervals use 10,000 paired
bootstrap samples of shared case IDs within source families, carrying each
sampled case across checkpoints and repeating the same aggregation.

\begin{figure}[!htbp]
\centering
\includegraphics[width=\textwidth]{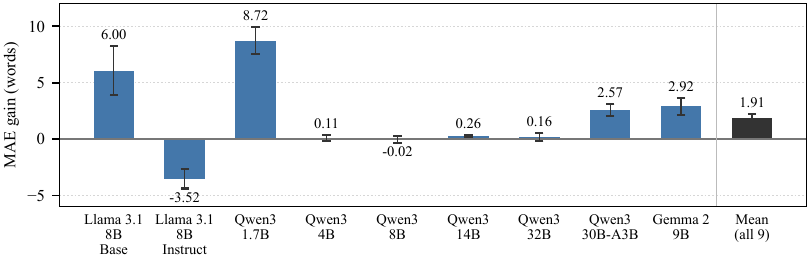}
\caption{Natural-trajectory MAE gains (shared minus model-specific) for all nine eligible checkpoints, with paired-bootstrap 95\% intervals. Discovery--confirmation curves appear in Figure~\ref{fig:policy-identification}.}
\label{fig:response-replication-all}
\end{figure}

Table~\ref{tab:natural-prediction} gives the per-checkpoint results and the
zero-edit baseline on exactly the same transitions. The zero-edit baseline
has lower point-estimate MAE than the model-specific curve for Llama Instruct,
Qwen3 1.7B, 8B, 14B, 32B, and 30B-A3B. No intervals for those baseline
contrasts are reported. A low prediction error can reflect small actual edits,
even when the requested correction remains unmet. The comparison with the
shared curve tests the predictive value of model identity, but does not
establish that the full curve outperforms a fitted model-specific bias or gain.

\begin{table}[!htbp]
\centering\footnotesize
\setlength{\tabcolsep}{3pt}
\begin{tabular*}{\linewidth}{@{\extracolsep{\fill}}lrrrr@{}}
\toprule
\textbf{Model} & \textbf{Zero edit} & \textbf{Shared} & \textbf{Specific} & $\mathrm{Gain}$ [\textbf{95\% CI}] \\
\midrule
Llama 3.1 8B Base & 86.23 & 88.67 & 82.67 & $+6.00\;[3.89,\,8.25]$ \\
Llama 3.1 8B Instruct & 8.00 & 6.26 & 9.78 & $-3.52\;[-4.39,\,-2.65]$ \\
Qwen3 1.7B & 5.68 & 14.57 & 5.85 & $+8.72\;[7.52,\,9.96]$ \\
Qwen3 4B & 12.10 & 11.27 & 11.16 & $+0.11\;[-0.19,\,0.40]$ \\
Qwen3 8B & 7.17 & 7.80 & 7.82 & $-0.02\;[-0.33,\,0.29]$ \\
Qwen3 14B & 4.28 & 5.83 & 5.58 & $+0.26\;[0.16,\,0.35]$ \\
Qwen3 32B & 3.79 & 5.08 & 4.92 & $+0.16\;[-0.20,\,0.52]$ \\
Qwen3 30B-A3B & 3.24 & 6.17 & 3.60 & $+2.57\;[2.07,\,3.07]$ \\
Gemma 2 9B & 15.30 & 11.41 & 8.49 & $+2.92\;[2.17,\,3.68]$ \\
\bottomrule
\end{tabular*}
\par\smallskip
$\mathrm{Gain}=\mathrm{Shared}-\mathrm{Specific}$; positive values favor model-specific prediction.
\caption{Prediction MAE (words) on independent natural revision trajectories for all nine eligible checkpoints. Intervals for the shared-minus-specific gain are paired, source-stratified case bootstrap intervals. Lower MAE is better.}
\label{tab:natural-prediction}
\label{tab:zero-edit-baseline}
\end{table}

Contraction means $|d_{t+1}|<|d_t|$. It can occur even when an edit crosses
the target, and net trajectory contraction does not require contraction at
every step.

The case is the inferential unit. Endpoint intervals use
source-stratified bootstrap; matched binary contrasts use exact McNemar tests
and Holm correction; recurrence models cluster by case. Cross-checks recount
stored text, recompute anchors, enforce zero split overlap, and require
byte-identical crossover drafts. Because greedy trajectories can differ across
backends, the Qwen scale pattern is reported under one uniform vLLM panel.

\paragraph{Conditional recoverability.}
The conditional-rescue analysis freezes revision 4 before API extension,
includes only trajectories still unresolved with at least one later revision,
and predicts capture through revision 8 from exact-state repeat fraction. The
reported odds ratio compares repeat fractions of 1 versus 0. Its adjustment set
contains current signed residual, absolute residual, log absolute residual,
target, missing anchors, minimum absolute residual, improvement from the initial to
current absolute residual, prior contractions, sign changes, no-op actions,
largest action, length-basin exit, checkpoint, source, and length band. Sandwich
covariance clusters identical case IDs across controllers; the retained Gemini
provider refusal is excluded under the endpoint audit rule.

Table~\ref{tab:recurrence-rescue-or} in the main text reports the adjusted associations.

\begin{figure}[!htbp]
  \centering
  \includegraphics[width=\textwidth]{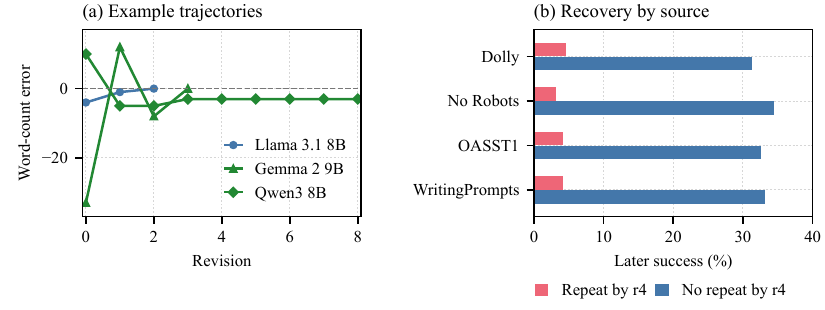}
  \caption{Matched exact-length trajectories show divergent controller
dynamics, while recurrence by revision 4 is associated with lower later
recoverability across all four sources.}
  \label{fig:main-law}
\end{figure}

\paragraph{Recurrence robustness and descriptive summaries.}
In the local panel, the unadjusted revision-4 association is negative in 11 of
12 checkpoints and all four sources and remains negative at revision-3 and
revision-5 landmarks. Qwen3 1.7B is the ceiling-limited exception: 470 exposed
and four unexposed cases yield a $+0.21$-point raw difference. In the API panel,
the adjusted coefficient remains negative at both adjacent landmarks, the raw
contrast is negative in all four sources, and both over-length and under-length
subsets preserve the negative combined association. The raw within-controller
direction is negative in four of seven API controllers. GPT, Gemini, and Claude
have only 3, 1, and 3 recurrence-exposed risk-set trajectories, respectively;
all four API controllers with at least 28 exposed cases are negative.

A preregistered early-capture/recurrence ratio ranks final success in both the
local ($\rho=0.783$) and API ($\rho=0.857$) panels. Early capture alone ranks
at least as strongly, and recurrence adds no held-family Brier-score improvement
\citep{brier1950verification}. We use the ratio as a descriptive controller
summary.

A length-basin exit is any observed transition from
$0<|d_t|\leq20$ to $|d_{t+1}|>50$ by the landmark. Observed stationary failure
means an unresolved trajectory whose final two outputs are identical, distinct
from the three-output persistent fixed-point criterion above; its
response-curve simulation analogue is unresolved capture with final predicted
action magnitude below 0.5 words.

\subsection{State-Matched History Intervention}
\label{app:history-reset}

The held-out first-recurrence screen produces 113 Llama and 194 GLM candidates.
The paired contrast varies retained dialogue as a block and does not separately
identify recency, position, or semantic content, which change together.
Applying the frozen revision-$\geq3$ rule excludes 28 Llama and 55 GLM states;
the discovery-selected polarity rule then excludes 2 non-over-length Llama and
19 non-under-length GLM states, leaving 83 and 120. For each retained state, the
two arms have the identical current draft, target, count, anchors, verifier
message, decoding configuration, and remaining revision budget. The full-history
arm preserves earlier turns; the reset arm preserves only the task, current
draft, and current verifier state. Removing earlier dialogue jointly changes
its content, length, turn structure, token positions, recency, and attention
context.

Estimands are paired state-level differences in first-step recurrence escape,
correct direction, absolute length-error contraction, later recurrence, final exact length,
all-anchor retention, and final $J_{\min}$. Confidence intervals use 20,000
paired bootstrap resamples within source while preserving observed source
sizes; they target the unweighted mean paired difference. Exact McNemar tests
target the same paired marginal but condition on discordant pairs (19 for
Llama final $J_{\min}$). The percentile interval and exact test differ in
construction, so its exclusion of zero can coexist with $p=0.063568$.
The frozen treatment gate requires a final-$J_{\min}$ gain of at least 10
points, a positive bootstrap lower bound, and a nonnegative point estimate for
all-anchor retention.

Figure~\ref{fig:interventions-combined} displays the exact-length paired
treatment effects with stratified paired-bootstrap 95\% intervals. Improvements in escape and final success
are separate intervention effects, not estimates of mediation through escape.
In absolute terms, Llama final joint success rises from 9/83 (10.8\%)
with full history to 18/83 (21.7\%) after reset; GLM rises from 4/120 (3.3\%)
to 10/120 (8.3\%). Thus 65/83 Llama and 110/120 GLM reset continuations
still fail within the remaining budget. These absolute endpoints, rather
than the magnitudes of the paired gains alone, establish that most reset
continuations remain unsuccessful.
The four partial-history conditions are categorical interventions on the
same states, not a continuous history-dose scale.

\subsection{Cross-Constraint History Effects}
\label{app:crossconstraint-history}

We compare full history and reset on 868 matched states from
Gemma 2 9B, Qwen3 14B, and Falcon-H1 7B. For each case, we select the first eligible repeated draft before
the trajectory ends. Selection is capped at 60 states for each
combination of model, task structure, and source. The tasks require specified
words at given word positions, specified words at sentence
ends, or per-sentence word counts within given bounds.
Word-position tasks also constrain total word count; the other two
constrain sentence count.

\begin{figure}[t]
\centering
\includegraphics[width=\textwidth]{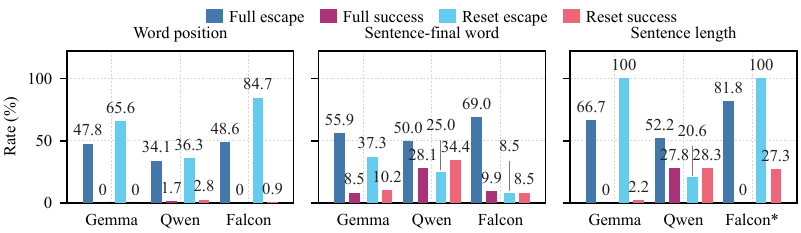}
\caption{Escape and final success with full or reset history at matched recurrent states. Sentence length: Gemma $n=45$; Falcon* $n=11$.}
\label{fig:history-crossconstraint}
\end{figure}

\textbf{History can help or hinder escape from repetition}, depending on
the model and task. Changes in escape need not be matched by changes in
final success (Figure~\ref{fig:history-crossconstraint}). For Gemma, escape increases by
17.8 percentage points on word-position tasks, with no change in final
success. On sentence-length tasks, escape rises by 33.3 points and
final success by 2.2 points. On sentence-final word tasks, escape instead falls
by 18.6 points, while final success rises by 1.7 points.
Final success is low in these Gemma comparisons. Qwen's sentence-length
task provides a comparison with higher success rates:
reset lowers escape by 31.7 points, while final success is 27.8\% with
full history and 28.3\% with reset. The reset-minus-full success difference is
0.6 points (95\% CI $-2.8$ to $3.9$;
Table~\ref{tab:crossconstraint-history-effects}).

\textbf{State selection also matters.} Some outputs already repeat at the first
revision; resetting these states reconstructs the earlier prompt that
produced the repeated output. Excluding these states reverses some escape
effects, including Qwen on sentence-length tasks
(Appendix~\ref{app:trigger-sensitivity}).

The targeted confirmation freezes 868 recurrent states and 1,736 paired arms:
284 Gemma 2 9B states, 391 Qwen3 14B states, and 193 Falcon-H1 7B states across
c05, c10, and c12. Both arms share the current task, draft, verifier report,
decoding, and remaining budget; only earlier dialogue is removed. Runtime
manifests, pair keys, state hashes, and deterministic verifier replay pass for
all arms. Falcon's fused-Mamba output was generated in disjoint resumable
shards and merged only after an exact expected-arm and duplicate audit.

Primary effects are within-state reset-minus-full-history differences.
Intervals use 20,000 constraint-by-source-stratified paired bootstrap draws.
Family summaries macro-average estimable constraint cells with equal cell
weight; the cross-family gate and c05/c12 controller-by-structure interaction were
frozen before outcome aggregation. The interaction uses HC3 covariance and
tests escape, contraction, and final success separately. Sparse cells are
excluded from formal pooled estimates.

\begin{table}[htbp]
\centering\footnotesize
\setlength{\tabcolsep}{3pt}
\begin{tabular}{llrrr}
\toprule
\textbf{Model} & \textbf{Task} & $n$ & \textbf{Escape} $\Delta$ & \textbf{Success} $\Delta$ [\textbf{95\%} \textbf{CI}] \\
\midrule
Gemma 2 9B & Word position & 180 & +17.8 & +0.0 [0.0, 0.0] \\
Gemma 2 9B & Sentence-final word & 59 & -18.6 & +1.7 [-3.4, 6.8] \\
Gemma 2 9B & Sentence length & 45 & +33.3 & +2.2 [0.0, 6.7] \\
Qwen3 14B & Word position & 179 & +2.2 & +1.1 [-1.7, 3.9] \\
Qwen3 14B & Sentence-final word & 32 & -25.0 & +6.2 [-9.4, 21.9] \\
Qwen3 14B & Sentence length & 180 & -31.7 & +0.6 [-2.8, 3.9] \\
Falcon-H1 7B & Word position & 111 & +36.0 & +0.9 [0.0, 2.7] \\
Falcon-H1 7B & Sentence-final word & 71 & -60.6 & -1.4 [-7.0, 2.8] \\
Falcon-H1 7B & Sentence length* & 11 & +18.2 & +27.3 [9.1, 45.5] \\
\bottomrule
\end{tabular}
\caption{Reset-minus-full differences (percentage points). Only final success has supplied paired 95\% intervals; the starred Falcon sentence-length cell is descriptive.}
\label{tab:crossconstraint-history-effects}
\end{table}

\FloatBarrier
\subsection{Partial-History Matched Intervention}
\label{app:partial-history}

The GLM pilot selects 60 states first observed in deep recurrence
(revision $\geq4$), balanced across c05, c10, and c12. Each state has four
continuations: full history, the last two draft--feedback pairs, the last pair,
and current task--draft--verifier state only. The 240 condition--state rows
share the identical current state, model, decoding, and remaining budget and
pass deterministic verifier replay.

The frozen primary family contains two paired escape contrasts, last-one minus
full-history and last-two minus full-history, tested by exact paired tests with
Holm correction. Confidence intervals use 20,000
constraint-by-source-stratified paired bootstrap draws and macro-average the
three structures equally. Contraction, later recurrence, and final success
are secondary endpoints. Table~\ref{tab:partial-history-arms} reports all four
conditions; Table~\ref{tab:partial-history-primary} gives the primary contrasts.

\begin{table}[!htbp]
\centering
\footnotesize
\setlength{\tabcolsep}{5pt}
\begin{tabular}{lrrrr}
\toprule
\textbf{History Retained} & \textbf{Escape} & \textbf{Contraction} & \textbf{Later Recurrence} & \textbf{Final Success} \\
\midrule
Full history & 10.0 & 8.3 & 86.7 & 0.0 \\
Last two pairs & 25.0 & 16.7 & 88.3 & 0.0 \\
Last pair & 51.7 & 25.0 & 76.7 & 3.3 \\
Reset & 66.7 & 25.0 & 60.0 & 5.0 \\
\bottomrule
\end{tabular}
\caption{GLM outcomes (\%) across four history conditions on the same 60 states, equally weighted across structures.}
\label{tab:partial-history-arms}
\end{table}

\begin{table}[h]
\centering
\footnotesize
\setlength{\tabcolsep}{6pt}
\begin{tabular}{lrrr}
\toprule
\textbf{Primary Escape Contrast} & \textbf{Difference (Points)} & \textbf{95\% CI} & \textbf{Holm} $p$ \\
\midrule
Last pair $-$ full history      & $+41.7$ & 30.0--53.3 & $1.19\times10^{-7}$ \\
Last two $-$ full history       & $+15.0$ & 3.3--26.7  & 0.0352 \\
\bottomrule
\end{tabular}
\caption{Frozen primary paired contrasts in the GLM partial-history pilot.
The unit is the recurrent state; values macro-average c05, c10, and c12.}
\label{tab:partial-history-primary}
\end{table}

\section{Aligned Checkpoint Localization}
\label{app:localization}

This supplementary study uses Llama as a worked example and includes
aligned Qwen3 8B and Gemma 2 9B checkpoints, with the
input interface held fixed within each pair. The studies test error readout,
Base-weight restoration into Instruct, activation transplantation, and
Instruct-weight grafts into Base. They do not assume that corresponding
layer ranges implement the same function across architectures.

\paragraph{Mechanistic localization.}
Causal tracing \citep{meng2022locating}, activation patching
\citep{zhang2023activationpatching}, and weight grafting
\citep{nief2025grafting} provide tools for locating information and
testing how model components affect behavior.
Work on subspace choice \citep{makelov2024subspace} and interactions among
mediators \citep{vaidyanathan2026mediators} shows how intervention design
affects causal attribution.
We use probes and interventions to examine error information and revision
behavior, with an aligned Llama Base--Instruct pair as the main case and
complementary Gemma and Qwen interventions.
We evaluate individual edits and complete revision trajectories separately
to determine whether changes in local behavior accompany gains in final
success.

\subsection{Operations and Readout Conventions}
\label{sec:checkpoint-boundary}
\paragraph{Blocks and parameter groups.}
A block is one decoder layer, indexed from zero: blocks 0--31 in Llama,
0--35 in Qwen, and 0--41 in Gemma. A block intervention selects every
checkpoint tensor under \texttt{model.layers.\{i\}.*}, including the
attention, MLP, and within-block normalization parameters. The output
package separately selects \texttt{model.norm.*} and \texttt{lm\_head.*}.
Unselected tensors retain the recipient checkpoint's values.

\paragraph{Probes.}
We read the representation at the final prompt token, before generating
an edit, from the model's returned \texttt{hidden\_states} tuple.
Probe labels L0, L1, \ldots\ index this tuple, whose first entry is the
embedding representation; they are not zero-based parameter-block labels.
The prespecified readouts are L32 for Llama, L36 for Qwen, and L42 for
Gemma. We standardize features using the training split and fit ridge
readouts for error sign and required-correction magnitude (2 versus 10
words). Regularization is selected by case-grouped cross-validation on
discovery data and frozen before confirmation. Readability does not
establish that generation uses the decoded information.

\paragraph{Restoration and grafting.}
Restoration uses Instruct as recipient and Base as donor. For each
selected tensor, half restoration sets
$\theta'=0.5\theta_{\mathrm{Instruct}}+0.5\theta_{\mathrm{Base}}$;
interpolation is computed in float32 and cast back to the recipient dtype.
Full restoration copies the Base tensor, while sham preserves Instruct
weights. Half restoration mixes every selected tensor, rather than
replacing half the selected blocks. Grafting reverses the direction:
selected Instruct tensors replace their counterparts in a Base recipient.
These are weight interventions, without additional training.

\paragraph{Activation transplantation.}
This operation instead replaces donor activations in a recipient forward
pass on identical tokenized inputs. The Llama experiments intervene on
post-block residuals over the feedback-message span during prompt prefill,
in both Base-to-Instruct and Instruct-to-Base directions; generated-token
states are untouched. The subsections below specify the sites, samples,
and behavioral criteria for each experiment.

\subsection{Study Coverage and Evaluation Splits}

For Qwen and Gemma, fixed-state discovery uses 24 cases (96 states), and
confirmation uses 48 disjoint cases (192 states). The 480-case closed-loop
comparisons include the discovery subset and are not wholly held out.
Activation and graft findings below are restricted to completed discovery
comparisons. A failed transplantation test does not establish the absence
of a mechanism outside the tested sites, spans, or intervention scheme.

\begin{table}[H]
\centering\footnotesize
\setlength{\tabcolsep}{4pt}
\begin{tabular}{>{\raggedright\arraybackslash}p{.67in}>{\raggedright\arraybackslash}p{1.35in}>{\raggedright\arraybackslash}p{1.28in}>{\raggedright\arraybackslash}p{1.65in}}
\toprule
\textbf{Pair} & \textbf{Probes and Restoration} & \textbf{Activation Transplantation} & \textbf{Weight Grafts} \\
\midrule
Llama & Held-out probes; restoration confirmed on 360 disjoint cases & Discovery complete; mediation criterion unmet & Discovery complete; direction improves, calibration criterion unmet \\
Qwen & Held-out probes and fixed-state dose confirmation; 480-case loop comparison & Discovery complete; no selected group passes & Discovery complete; no selected group passes \\
Gemma & Held-out probes and fixed-state dose confirmation; 480-case loop comparison & Discovery complete; no selected group passes & Partial transfer in discovery; confirmation incomplete \\
\bottomrule
\end{tabular}
\caption{Coverage and confirmation status of the aligned mechanism studies. Passing criteria are study-specific.}
\label{tab:mechanism-coverage}
\end{table}

\subsection{Llama Residual Probes}
\label{app:llama-residual-probes}
Meta Llama 3.1 8B Base and Instruct receive byte- and token-identical Meta
serialization on 288 fixed feedback states from 72 cases. Each case supplies
residuals $d\in\{-10,-2,+2,+10\}$; 24 cases form discovery and 48 disjoint
cases form confirmation. Representations are captured from the returned hidden-state tuple at
the final prompt token, using the readout convention defined above. Ridge regularization is selected separately for each
checkpoint and readout by six-fold GroupKFold over discovery cases at layer 32
and then frozen across layers; confirmation data are not used to select
regularization. Layer-32 estimates are prespecified; the layer sweep below is descriptive.
Primary readouts are layer-32 error-sign accuracy and required-correction magnitude
classification (2 versus 10 words); best-layer values are descriptive.

\begin{table}[!htbp]
\centering\footnotesize
\setlength{\tabcolsep}{4pt}
\begin{tabular}{llrr}
\toprule
\textbf{Model} & \textbf{Checkpoint} & \textbf{Error Sign} & \textbf{Required Magnitude} \\
\midrule
Llama 3.1 8B & Base & 90.1 & 62.5 \\
Llama 3.1 8B & Instruct & 88.0 & 85.4 \\
Qwen3 8B & Base & 80.7 & 80.7 \\
Qwen3 8B & Instruct & 88.0 & 80.7 \\
Gemma 2 9B & Base & 94.3 & 69.3 \\
Gemma 2 9B & Instruct & 92.2 & 65.6 \\
\bottomrule
\end{tabular}
\caption{Prespecified final-layer probe accuracy (\%) on 192 held-out states per checkpoint. Magnitude distinguishes 2 from 10 words.}
\label{tab:mechanism-probes}
\end{table}

The probe table reports the prespecified final hidden-state readout: L32 for
Llama, L36 for Qwen, and L42 for Gemma. All use the final prompt token;
required-correction magnitude refers to the requested 2- versus 10-word
change, not the magnitude of the generated edit. Readout accuracy does not
by itself show that generation uses the decoded information.

A descriptive layer sweep shows the output-layer correction-magnitude deficit is not
an absence of the underlying quantity: $R^2$ rises from $0.08$ at layer 32 to
$0.62$ for Base at layer 14, and from $0.44$ to $0.93$ for Instruct at layer 8.
These layers are selected on the split used to report them, so the values are
upward-biased and not comparable to the preregistered layer-32 estimates.

\subsection{Llama Parameter Restoration}
\begin{figure}[!htbp]
\centering
\includegraphics[width=\textwidth]{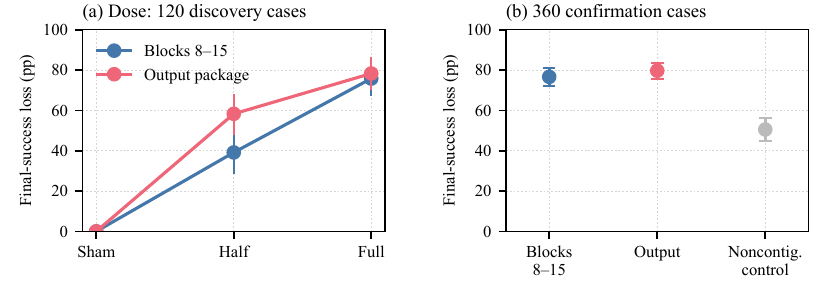}
\caption{Base-weight restoration: discovery dose response (a) and disjoint confirmation (b). Bars show paired-bootstrap 95\% intervals.}
\label{fig:llama-restoration}
\end{figure}

Base weights are restored into an otherwise fixed Instruct checkpoint for four
contiguous eight-block bands, final norm plus LM head, a sham copy, and an
eight-block noncontiguous parameter-count control. The discovery screen uses
96 fixed states and Main-120. Selected blocks 8--15 and the output package are
then evaluated at sham, half, and full restoration and confirmed on 360 cases
absent from discovery. Fixed-state intervals resample 24 case clusters;
closed-loop contrasts are paired by case and source. Figure~\ref{fig:llama-restoration} reports final-success losses relative to
sham with source-stratified paired-bootstrap 95\% intervals. These tests establish
necessity under the intervention; unique contiguity remains unresolved.

\subsection{Llama Activation Transplantation}
On the 96 discovery states, donor and recipient share identical token IDs.
During prompt prefill only, the complete feedback-message span is transplanted
in both directions either across post-block residuals 8--15 or at the
post-block-31 output boundary; generated-token states are untouched. For each
unit and residual family (magnitude 10 correction or magnitude 2 near-target),
we report the clean checkpoint contrast, Instruct-to-Base denoising effect,
Base-to-Instruct noising loss, and their interaction on direction and clipped
calibration utility. Instruct-to-Base transplantation raises correct revision
direction by 20.8--29.2 points while worsening magnitude calibration. No unit
passes the frozen bidirectional mediation gate.

\subsection{Llama Controller Graft}

Five Base-recipient conditions share the Meta interface and greedy decoding:
sham, blocks 8--15, output package, their joint graft, and a matched
noncontiguous-plus-output control. Each runs 96 fixed states and Main-120;
intervals resample case or source strata. The joint graft improves correct editing direction but fails the
prespecified calibration criterion. Calibration utility is
$-\min(|a/c-1|,10)$ for actual word-count change $a$ and requested
change $c\ne0$. The gate requires nonnegative mean utility differences
against both Base and the noncontiguous-plus-output control; the observed
differences are $-0.268$ and $-2.199$, respectively. The direction gain
therefore does not meet the criterion for transferring effective revision
behavior.

\begin{figure}[!htbp]
\centering
\includegraphics[width=\textwidth]{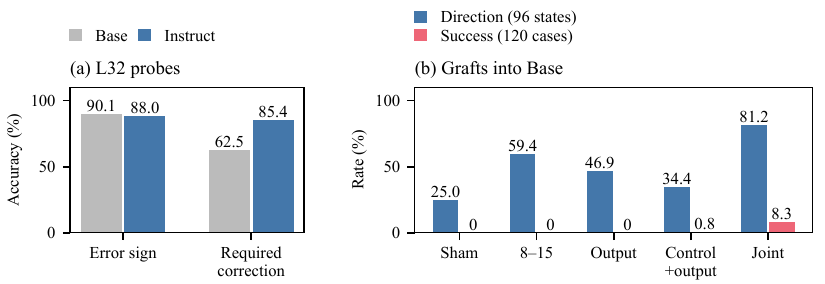}
\caption{Llama probe readouts and all graft conditions. Direction and final success use separate fixed-state and closed-loop samples.}
\label{fig:llama-full-localization}
\end{figure}

On the 96 fixed states, direction accuracy is 25.0\% for Base,
59.4\% with the block graft, 46.9\% with the output graft, and 81.25\%
with their joint graft. The joint graft succeeds on 10 of 120 separate
closed-loop cases (Figure~\ref{fig:llama-full-localization}).

\subsection{Qwen: Restoration Effects Without Successful Transfer}
\label{app:qwen-localization}
Qwen3 8B Base and Instruct share a 36-block architecture and the same
Instruct serialization. Tokenizer and tensor-shape audits pass. Probe
regularization is selected using discovery cases and frozen before
confirmation (Table~\ref{tab:mechanism-probes}). Error direction remains
readable in both checkpoints. Error expansion denotes an increase in
absolute word-count error after an edit, $|d_{t+1}|>|d_t|$. The stored
fields call this outcome \texttt{overshoot} (or its complement
\texttt{overshoot\_avoidance}); it need not involve crossing the target.

The restoration screen tests four nine-block bands, the output package,
a sham, and a parameter-count-matched noncontiguous control. Blocks 0--8
and 9--17 are selected in discovery. On the 192 held-out fixed states,
full restoration increases error expansion by 62.0 and 21.4 percentage points,
respectively; paired 95\% intervals are [50.0, 72.9] and [10.9, 32.8].
Half-restoration effects are weaker. Both groups satisfy the frozen
restoration criteria, including comparison with the noncontiguous control.
The 480-case closed-loop comparisons show lower success and larger
terminal error (Table~\ref{tab:mechanism-restoration}).

Bidirectional activation transplantation covers 384 patched discovery
cells across the two selected groups. Neither group passes the required
behavioral-direction and calibration criteria. Seven graft conditions
compare Base, output only, noncontiguous blocks plus output, and each
selected block group with and without output. Neither joint graft passes
the selected behavioral-signature criteria against Base and the control.
On fixed discovery states, the two joint grafts increase error expansion
relative to Base by 28.1 and 31.3 percentage points, respectively.
No activation or graft confirmation split is opened. Thus, the restoration
results establish sensitivity to these weight changes, without showing
that the tested components transfer effective revision behavior.

\subsection{Gemma: Different Effects on Repetition and Correction}
\label{app:gemma-localization}
Gemma 2 9B Base and Instruct use a common interface and matched tokenized
fixed states. The discovery restoration screen selects blocks 21--30
and 31--41 for half- and full-dose comparisons with sham and a
parameter-count-matched noncontiguous control. Both groups meet the
prespecified criteria on the 192 held-out fixed states. Full restoration
of blocks 21--30 lowers correct direction by 65.6 percentage points
[57.8, 73.4], whereas restoring blocks 31--41 increases error expansion by
33.9 points [23.4, 44.3]. These interventions therefore produce different
forms of poor revision, rather than a uniform loss of activity.

\begin{figure}[htbp]
\centering
\includegraphics[width=0.72\textwidth]{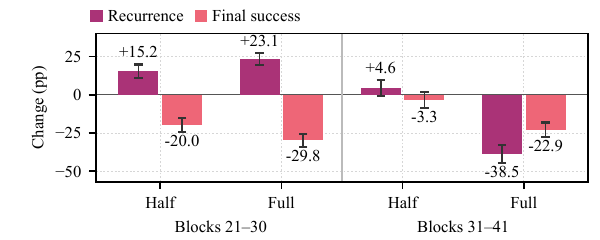}
\caption{Gemma Base-weight restoration into Instruct. Changes relative to sham in recurrence and final success across 480 cases, with source-stratified paired-bootstrap 95\% intervals. Half and full denote interpolation weights 0.5 and 1.}
\label{fig:gemma-restoration}
\end{figure}

\begin{table}[!htbp]
\centering\footnotesize
\setlength{\tabcolsep}{4pt}
\begin{tabular}{llrrr}
\toprule
\textbf{Model} & \textbf{Blocks} & \textbf{Repeat} $\Delta$ (\textbf{pp}) & \textbf{Success} $\Delta$ (\textbf{pp}) & \textbf{Error} $\Delta$ (\textbf{words}) \\
\midrule
Qwen3 8B & 0--8 & -5.4 [-9.0, -1.9] & -6.5 [-9.0, -4.0] & +77.9 [+71.2, +85.0] \\
Qwen3 8B & 9--17 & +3.5 [+1.0, +6.0] & -5.2 [-7.9, -2.5] & +15.6 [+11.6, +20.1] \\
Gemma 2 9B & 21--30 & +23.1 [+19.2, +27.3] & -29.8 [-34.2, -25.4] & +80.5 [+70.7, +90.6] \\
Gemma 2 9B & 31--41 & -38.5 [-44.4, -32.7] & -22.9 [-27.7, -18.1] & +118.8 [+107.6, +130.6] \\
\bottomrule
\end{tabular}
\caption{Full restoration minus sham on 480 cases per model, with source-stratified paired-bootstrap 95\% intervals.}
\label{tab:mechanism-restoration}
\end{table}

Here recurrence means any exact repeated output among the recorded drafts,
including the initial draft, across all 480 cases; it is not conditioned
on final failure as in Figure~\ref{fig:failure-anatomy}.
Restoring blocks 21--30 increases recurrence by 23.1 points and lowers
success by 29.8 points. Restoring blocks 31--41 instead reduces recurrence
by 38.5 points while lowering success by 22.9 points and increasing
terminal absolute error by 118.8 words. The latter success rates are
32.5\% for sham and 9.6\% after restoration. These are joint consequences
of the weight intervention; they do not identify a causal effect of
recurrence on success. Broad disruption from weight replacement also
limits anatomical interpretation.

Neither selected group passes the bidirectional activation-transplantation
criteria on 384 discovery cells. Grafting Instruct blocks 21--30 plus
output into Base does improve direction from 51.0\% to 74.0\% on 96
fixed states, and success from 1/120 to 11/120 on separate discovery
trajectories (Table~\ref{tab:gemma-graft}). Calibration improves relative
to Base but remains below the noncontiguous-plus-output control.

\begin{table}[!htbp]
\centering\footnotesize
\setlength{\tabcolsep}{4pt}
\begin{tabular}{lrrr}
\toprule
\textbf{Condition} & \textbf{Direction} (\%) & \textbf{Calibration Utility} & \textbf{Final Successes} \\
\midrule
Base & 51.0 & -8.99 & 1/120 \\
Output & 50.0 & -8.82 & 0/120 \\
Noncontiguous + output & 63.5 & -3.54 & 3/120 \\
Blocks 21--30 & 75.0 & -5.46 & 8/120 \\
Blocks 21--30 + output & 74.0 & -5.13 & 11/120 \\
Blocks 31--41 & 36.5 & -3.18 & 2/120 \\
Blocks 31--41 + output & 34.4 & -3.21 & 1/120 \\
\bottomrule
\end{tabular}
\caption{Gemma graft discovery: 96 fixed states and 120 separate closed-loop cases. Higher calibration utility is better.}
\label{tab:gemma-graft}
\end{table}

The Gemma discovery rule requires improvement on a selected behavior
relative to Base and the control, with no contradiction on other selected
behaviors relative to Base. Unlike Llama's calibration rule, it does not
require calibration to be at least as good as the control; discovery also
does not require an interval excluding zero. Blocks 21--30 pass this
screen, whereas blocks 31--41 fail because direction deteriorates.
The subsequent graft confirmation is incomplete and is excluded from
all reported estimates and conclusions. We therefore describe the
21--30 result as discovery-only partial transfer, not confirmed recovery.

\section{Aligned Post-Training and Interface Comparisons}
\label{app:tulu-stage-interface}

The Llama comparison in Appendix~\ref{app:llama-residual-probes}
uses Base and Meta Instruct under byte-identical Meta serialization.  As a
complementary behavioral comparison, we evaluate five released Llama--T\"ulu
checkpoints \citep{lambert2024tulu} under a common plain-text interface held fixed
across this aligned panel.  The four chat-capable checkpoints are also evaluated with their own
bundled native templates; Base has no native chat template and is excluded
from that comparison.  Cases are shared across checkpoints within each constraint family,
while the three constraint families retain their distinct datasets.  Table~\ref{tab:all-model-assay-matrix}
retains the canonical controller panel; this appendix reports a separately
frozen aligned-checkpoint ablation.

\begin{table}[h]
\centering
\footnotesize
\setlength{\tabcolsep}{5.2pt}
\begin{tabular}{llrrr}
\toprule
\textbf{Checkpoint} & \textbf{Interface} & \textbf{Exact Length} & \textbf{Lexical} & \textbf{Compositional} \\
\midrule
Llama 3.1 8B Base     & Common & 0.6 & 22.1 & 0.4 \\
T\"ulu 3 8B SFT       & Common & 4.4 & 21.9 & 29.0 \\
                       & Native & 4.6 & 22.9 & 33.1 \\
T\"ulu 3 8B DPO       & Common & 15.0 & 25.2 & 42.5 \\
                       & Native & 15.4 & 24.0 & 64.6 \\
T\"ulu 3.1 8B RLVR    & Common & 12.5 & 28.5 & 45.6 \\
                       & Native & 16.7 & 25.8 & 48.8 \\
Llama 3.1 8B Instruct & Common & 47.7 & 46.5 & 91.0 \\
                       & Native & 79.0 & 50.8 & 97.3 \\
\bottomrule
\end{tabular}
\caption{Final joint success (\%) across aligned checkpoints under common and
native interfaces.}
\label{tab:tulu-stage-interface}
\end{table}

Structure-level rates explain the strong interaction between checkpoint and constraint family under the common interface ($\chi^2(8)=1094.58$, $p=5.66\times10^{-231}$).  Base's 22.1\%
lexical average comes entirely from c07 (88.3\%), with 0\% on c05 and c12.
SFT leaves exact length and the lexical average nearly unchanged, yet reaches
55.4\% on c09 and only 2.5\% on c10.  DPO reaches 26.3\% on c10 and 15.0\%
on exact length; RLVR is higher on c12 and c10 but lower than DPO on exact
length and c09. Performance across the released checkpoints therefore changes differently
for different task structures (definitions are in
Appendix~\ref{app:verifiers}).

The effect of native formatting also varies across task structures.  SFT's modest $+4.2$-point
compositional change combines a 13.3-point decrease on c09 with a 21.7-point
increase on c10, whereas DPO's $+22.1$-point gain combines increases of 13.8
and 30.4 points.  This changes the DPO--RLVR ordering: RLVR leads by 3.1 points
under the common interface, but DPO leads by 15.8 under native serialization,
a paired difference-in-differences of 19.0 points (95\% bootstrap CI
12.9--25.2).  Meta Instruct has a different signature, gaining 31.2 points on
exact length (26.0--36.3).  Interface-by-checkpoint tests reject a uniform
template bonus for exact length ($\chi^2(3)=112.72$, $p=2.85\times10^{-24}$),
lexical constraints ($\chi^2(3)=22.40$, $p=5.40\times10^{-5}$), and
compositional constraints ($\chi^2(3)=46.03$, $p=5.58\times10^{-10}$).
The common interface compares checkpoints under the same formatting;
native templates measure their deployed behavior. Neither comparison
isolates the causal effect of a training stage.

\section{Additional Revision Diagnostics}
\subsection{Fixed-Draft Crossover}
\label{app:crossover}
The crossover uses 240 common cases, with 60 from each source. Three
Llama--peer pairs form ten unique draft-generator--reviser combinations: four diagonal
cells and six off-diagonal cells. All 2,400 trajectories pass task, anchor,
target, initial-text, and endpoint recount checks. The six within-draft
reviser contrasts are tested by exact McNemar tests with Holm correction.
Table~\ref{tab:fixed-draft} groups final-success rates by draft source
and identifies each reviser explicitly. Draft-source effects are
not uniformly zero: two point estimates favor peer drafts by 8--9 points,
but no draft-source contrast survives six-test Holm correction at .05.
Table~\ref{tab:crossover-conditional} reports the six paired contrasts
and the analysis conditional on initial failure. Excluding initially
successful drafts leaves 225 cases for Llama drafts, 234 for Qwen drafts,
and 240 each for Gemma and GLM drafts. The conditional advantage ranges
from 45.3 to 77.1 percentage points. Each excluded case succeeds initially
under both revisers because the draft is identical, so exclusion changes
the denominator but not the discordant-pair counts or exact test results.
The paired intervals in the all-case panel pertain only to that panel;
the conditional panel reports its frozen rates, differences, and tests.
\begin{table}[!htbp]
\centering
\setlength{\tabcolsep}{3pt}
\begin{tabular*}{\linewidth}{@{\extracolsep{\fill}}llrrrrr@{}}
\toprule
Draft source & Peer & $n$ & Llama (\%) & Peer (\%) & Difference (pp) & Holm $p$ \\
\midrule
\multicolumn{7}{l}{\textit{All cases; differences include paired-bootstrap 95\% intervals}} \\
Llama & Qwen & 240 & 76.2 & 17.1 & 59.2 [52.5, 65.8] & $4.59\times10^{-37}$ \\
Qwen & Qwen & 240 & 84.6 & 15.8 & 68.8 [62.5, 75.0] & $2.78\times10^{-45}$ \\
Llama & Gemma & 240 & 76.2 & 33.8 & 42.5 [35.0, 50.0] & $5.10\times10^{-22}$ \\
Gemma & Gemma & 240 & 81.2 & 25.4 & 55.8 [49.2, 62.5] & $1.65\times10^{-37}$ \\
Llama & GLM & 240 & 76.2 & 12.9 & 63.3 [56.7, 69.6] & $1.23\times10^{-39}$ \\
GLM & GLM & 240 & 85.4 & 8.3 & 77.1 [71.2, 82.5] & $4.44\times10^{-51}$ \\
\midrule
\multicolumn{7}{l}{\textit{Initially unsuccessful drafts only; conditional point estimates}} \\
Llama & Qwen & 225 & 74.7 & 11.6 & 63.1 & $4.59\times10^{-37}$ \\
Qwen & Qwen & 234 & 84.2 & 13.7 & 70.5 & $2.78\times10^{-45}$ \\
Llama & Gemma & 225 & 74.7 & 29.3 & 45.3 & $5.10\times10^{-22}$ \\
Gemma & Gemma & 240 & 81.2 & 25.4 & 55.8 & $1.65\times10^{-37}$ \\
Llama & GLM & 225 & 74.7 & 7.1 & 67.6 & $1.23\times10^{-39}$ \\
GLM & GLM & 240 & 85.4 & 8.3 & 77.1 & $4.44\times10^{-51}$ \\
\bottomrule
\end{tabular*}
\caption{Fixed-draft reviser contrasts. Llama: 3.1 8B; Qwen: Qwen3 14B; Gemma: 2 9B; GLM: GLM-4 9B. Differences are Llama minus peer. Frozen all-case intervals use 20,000 case-paired bootstrap resamples. Intervals apply only to the all-case panel. Exact McNemar tests are Holm-corrected over six comparisons separately in each panel.}
\label{tab:crossover-conditional}
\end{table}

The crossover changes the deployed reviser and native interface jointly.

\subsection{Capture and Budget}
\label{app:revision-dynamics}
The API trajectory audit contains 3,359 protocol-complete trajectories:
479 Gemini records and 480 for each other system. The budget replay uses
479 all-system-common cases instead. Conditional capture is computed among
trajectories unresolved before each revision. Its mean over revisions 5--6
is no lower than over 1--2 for every system, but the late means range from
6.4\% to 58.7\%. A low fixed-budget endpoint need not imply a decreasing
per-round capture probability.
\begin{figure}[!htb]
\centering
\includegraphics[width=\textwidth]{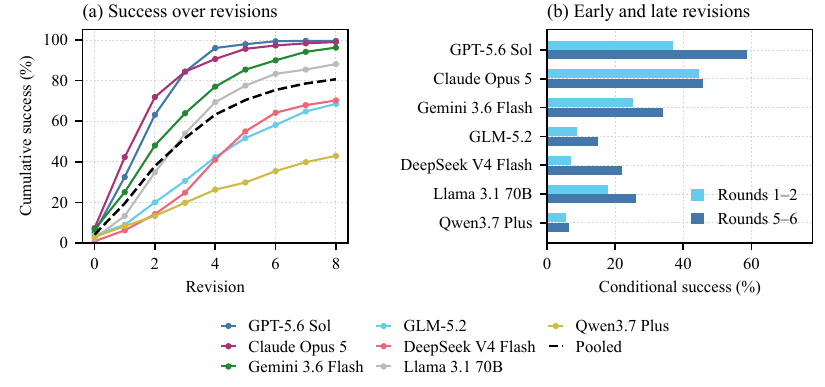}
\caption{Cumulative capture and conditional capture under the eight-revision budget on the protocol-complete API panel.}
\label{fig:capture-dynamics}
\end{figure}
Among the 650 failures, 90.2\% end at their own best absolute residual.
This does not rule out intermediate oscillation or instability. Five
trajectories attain exact word count and later leave it; this is not loss
of joint success. The exact-length diagnostic labels follow the priority rule in
Appendix~\ref{app:failure-categories}.
Figure~\ref{fig:additional-revision-diagnostics} retains the dense Qwen scale
comparison and residual-distance diagnostic. The latter uses disjoint integer
bins; zero denotes a best-at-terminal trajectory, not stable dynamics.
\begin{figure}[!htb]
\centering
\includegraphics[width=\textwidth]{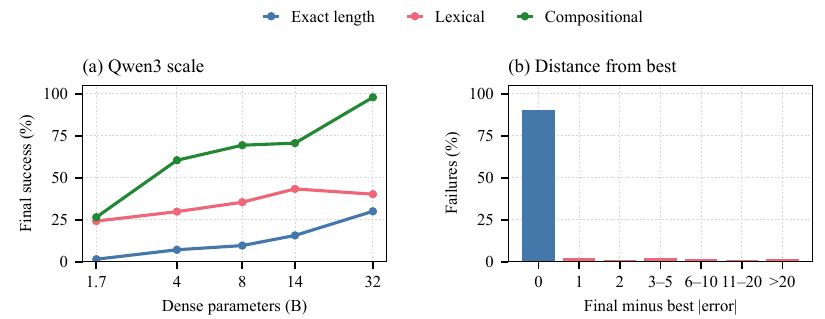}
\caption{Additional diagnostics: dense Qwen scale across constraint families and final distance from each failed trajectory's best residual.}
\label{fig:additional-revision-diagnostics}
\end{figure}

\subsection{Trigger-State Sensitivity of History Reset}
\label{app:trigger-sensitivity}
All 209 first-revision reset states replay the previously failing prompt;
199 repeat the trigger text. For a first-step fixed point, the current
draft and feedback equal the initial draft and feedback, so deleting the
history reconstructs the earlier input. This protocol property contributes
to mixed-cohort contrasts. Table~\ref{tab:trigger-sensitivity} reports all
controller--structure cells, including small cells, rather than selecting
only sign reversals. The subset analysis is retrospective and does not
replace the frozen full-cohort estimand. Removing first-revision states
changes the population; it is not a causal effect of intervention timing.

\begin{table}[ht]
\centering\footnotesize
\setlength{\tabcolsep}{3pt}
\begin{tabular}{llrrrr}
\toprule
\textbf{Model} & \textbf{Task} & $n$ & \textbf{Escape} $\Delta$ & \textbf{Later-state} $n$ & \textbf{Escape} $\Delta$ \\
\midrule
Gemma & Word position & 180 & +17.8 & 162 & +25.9 \\
Gemma & Sentence-final word & 59 & -18.6 & 48 & -4.2 \\
Gemma & Sentence length & 45 & +33.3 & 45 & +33.3 \\
Qwen & Word position & 179 & +2.2 & 150 & +12.7 \\
Qwen & Sentence-final word & 32 & -25.0 & 18 & +5.6 \\
Qwen & Sentence length & 180 & -31.7 & 92 & +13.0 \\
Falcon & Word position & 111 & +36.0 & 109 & +37.6 \\
Falcon & Sentence-final word & 71 & -60.6 & 24 & -16.7 \\
Falcon & Sentence length & 11 & +18.2 & 11 & +18.2 \\
\bottomrule
\end{tabular}
\caption{Reset-minus-full escape effects (percentage points), in all states and after excluding first-revision triggers.}
\label{tab:trigger-sensitivity}
\end{table}
For Qwen c10, the later-state subset has 92 states: escape changes from
19.57\% to 32.61\%, and final success from 10.87\% to 14.13\%.
The final-success difference is 3.26 points, with a descriptive 95\% interval
of $[-3.26,9.78]$, recomputed using 5,000 source-stratified paired bootstrap
draws (seed 20260908). The original cell-rate CSV supplies no escape
intervals; no interval is imputed from its marginal rates.

\section{Failure Trajectories across Constraint Families}
\label{app:all-constraint-failures}
We classify unsuccessful protocol-complete trajectories separately for exact
length, lexical constraints, and compositional constraints. All 19 model
configurations are included. There are 27,359 complete trajectories and
12,170 failures: 5,272 exact-length, 4,805 lexical, and 2,093 compositional
failures. One protocol-incomplete Gemini exact-length record is excluded.
These descriptive counts use all complete records, rather than restricting
every API model to the 479 cases shared for paired endpoint comparisons.

\paragraph{A common classification rule.}
Let $y_0,\ldots,y_K$ be the recorded drafts, including the initial draft,
where $K$ is the last recorded revision index. A trajectory is a failure
when its final draft does not satisfy all constraints for its task. Text
equality means exact UTF-8 equality without normalization; equal word counts
or equal verifier residuals do not establish equal outputs. Among failed
trajectories, assign the first matching category below:
\begin{enumerate}
\item \textbf{Terminal fixed point.} At least three drafts are recorded and
$y_K=y_{K-1}=y_{K-2}$. Earlier drafts need not be identical. This is an
observed terminal pattern, not a claim that repetition would continue forever.
\item \textbf{Terminal cycle.} For at least one $p\in\{2,3,4\}$, the last
$2p$ drafts consist of two identical consecutive blocks of $p$ drafts, each
containing at least two distinct outputs. For example, a trajectory ending
in $A,B,A,B$ satisfies the period-2 test when $A\ne B$. We do not require
the cycle to begin at the initial draft or claim to identify its minimal
period. The fixed-point category takes priority if both conditions hold.
\item \textbf{Other repeat.} Some $y_i=y_j$ with $i<j$, but
neither terminal condition holds. For example, $A,B,A,C$ contains repetition
without a terminal fixed point or tested cycle. A trajectory can receive this
label even if its final output differs from every earlier one.
\item \textbf{No repeat.} All recorded drafts are distinct.
This label describes text novelty only: it does not imply error reduction,
correct editing direction, or proximity to satisfying the constraints.
\end{enumerate}
These four categories are mutually exclusive and exhaustive under the stated
priority rule. They describe output trajectories, not distinct causal
mechanisms. Unlike the seven exact-length categories in
Appendix~\ref{app:failure-categories}, they do not use word-count thresholds,
anchor status, or signed-error flips. Consequently, a repeated trajectory
ending at exact word count with missing anchors remains in a repetition
category here, although the seven-category classifier assigns anchor failure.
The two classifications should not be compared as if their labels were
interchangeable.

\begin{figure}[!htbp]
\centering
\includegraphics[width=0.92\textwidth]{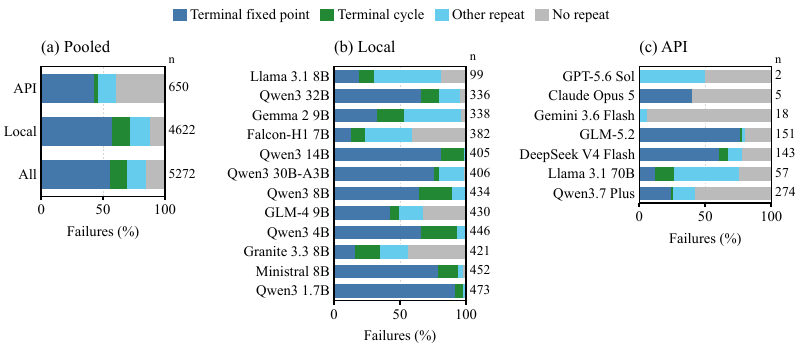}
\caption{Exact-length failures: API, local, and pooled compositions (a), and all 19 models (b,c). Colors and within-panel model order match Figure~\ref{fig:failure-anatomy} and Table~\ref{tab:all-model-assay-matrix}.}
\label{fig:failure-exact-appendix}
\par\vspace{5pt}
\includegraphics[width=0.92\textwidth]{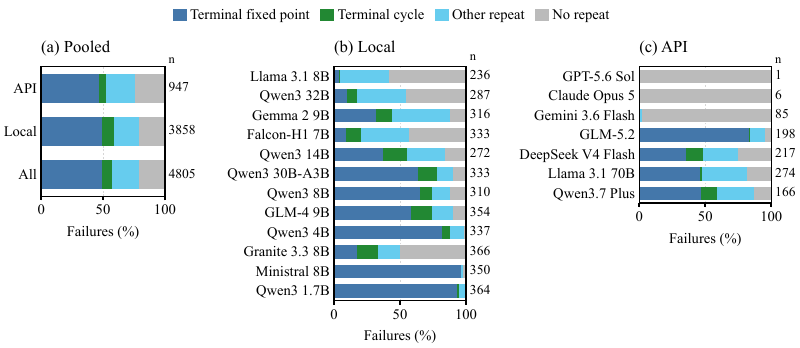}
\caption{Lexical failures: pooled compositions (a) and all 19 models (b,c); colors and model order follow Figure~\ref{fig:failure-exact-appendix}.}
\label{fig:failure-lexical-appendix}
\par\vspace{5pt}
\includegraphics[width=0.92\textwidth]{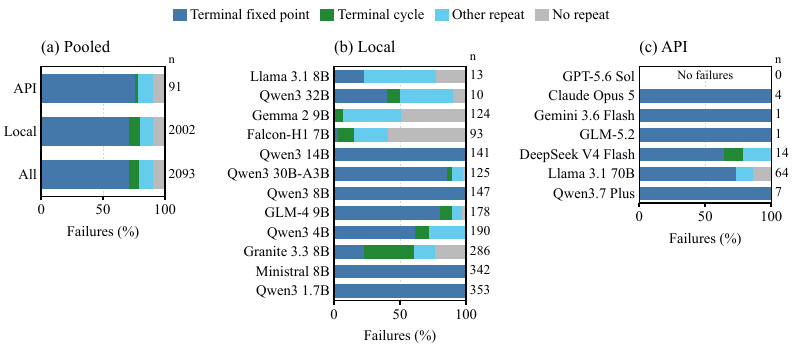}
\caption{Compositional failures: pooled compositions (a) and all 19 models (b,c); colors and model order follow Figure~\ref{fig:failure-exact-appendix}.}
\label{fig:failure-comp-appendix}
\end{figure}

\paragraph{Reconstruction and aggregation.}
Local labels are reconstructed from the existing frozen per-trajectory
indicators for fixed points, terminal cycles, and any output recurrence.
API labels are recomputed from the original recorded drafts after checking
input hashes against the existing analysis manifest. Each model contributes
480 complete trajectories per constraint family, except Gemini exact length
with 479. Lexical summaries preserve the original 240/120/120 allocation to
c05/c07/c12; compositional summaries preserve the 240/240 allocation to
c09/c10. Figures~\ref{fig:failure-exact-appendix}--\ref{fig:failure-comp-appendix}
show within-family compositions. Each bar is normalized by that model's
failure count, marked $n$; an empty bar marked ``No failures'' has no defined
failure composition. Panel (a) pools failed trajectories within API, local, or
all models, so models with more failures contribute more. They are not
averages that weight models or constraint structures equally.

\begin{table}[htbp]
\centering\footnotesize
\begin{tabular}{lrrrrr}
\toprule
\textbf{Constraint Family} & \textbf{Terminal Fixed Point} & \textbf{Terminal Cycle} & \textbf{Other Repeat} & \textbf{No Repeat} & \textbf{Total} \\
\midrule
Exact length & 2,922 & 715 & 833 & 802 & 5,272 \\
Lexical constraints & 2,348 & 418 & 1,017 & 1,022 & 4,805 \\
Compositional constraints & 1,491 & 172 & 225 & 205 & 2,093 \\
\bottomrule
\end{tabular}
\caption{Failure counts under the common output-trajectory classification.}
\label{tab:failure-common-counts}
\end{table}

\FloatBarrier

\end{document}